\documentclass{article} 
\usepackage{styles/iclr2027_conference,times}

\usepackage{amsmath,amsfonts,bm}

\def\eqref#1{equation~\ref{#1}}

\def\1{\bm{1}}

\DeclareMathAlphabet{\mathsfit}{\encodingdefault}{\sfdefault}{m}{sl}
\SetMathAlphabet{\mathsfit}{bold}{\encodingdefault}{\sfdefault}{bx}{n}

\usepackage{hyperref}
\usepackage{url}

\usepackage{graphicx}
\usepackage{booktabs}
\usepackage{multirow}
\usepackage{multicol}
\usepackage{wrapfig}
\title{Explainability from Training \\   with~Applications to TCR-Epitope Prediction}

\author{$^1$Jiarui Li, $^1$Zixiang Yin, $^2$Samuel J. Landry, $^1$Zhengming Ding, $^{1*}$Ramgopal R. Mettu \\
  $^1$ Department of Computer Science, Tulane University\\$^2$ Department of Biochemistry and Molecular Biology, Tulane University School of Medicine \\
  \texttt{\{jli78, zyin, landry, zding1, rmettu\}@tulane.edu} \\
  $^*$Corresponding author.\\
}

\usepackage{xstring}
\newcommand{\ModelName}[1][]{%
    \IfStrEqCase{#1}{%
        {norm}{explainability from training}%
        {cap}{Explainability from Training}%
        {first}{Explainability from training}%
        {highlight}{\underline{E}xplainability \underline{F}rom \underline{T}raining}%
        {abbr}{EFT}%
    }[\textbf{Error: invalid option for \textbackslash ModelName}]%
}

\iclrfinalcopy 
\begin{document}
    \maketitle
    \begin{abstract}
Deep learning models have achieved strong performance in artificial intelligence for science, yet their black-box nature limits our understanding of how they learn scientific tasks. Existing methods for interpretability provide limited insight into how models organize evidence and evolve during learning. We introduce \ModelName[highlight] (\ModelName[abbr]), a model-agnostic paradigm that traces model interpretation during training to explain why models rely on specific features and how they organize these features as predictive evidence.
We apply \ModelName[abbr] to four state-of-the-art T cell receptor (TCR)-epitope prediction models, TCR-SRIM, TULIP, MixTCRpred, and NetTCR-2.2, spanning post-hoc and interpret-by-design approaches as well as transformers and CNNs. To investigate how structural information affects model explanations, we introduce a benchmark, TCR-XAI2, containing 388 unique experimentally resolved TCR-epitope structures, complemented by structures predicted using AlphaFold3, Boltz-2, TCRModel2, tFold-TCR, and OpenFold3. Using \ModelName[abbr] with TCR-XAI2, we demonstrate that (1) CNN and transformer models exhibit distinct learning trajectories; (2) TCR $\alpha$ and $\beta$ evidence can conflict during learning, limiting the benefits of jointly modeling both chains, while MHC information mitigates this; and (3) real versus predicted structural data for TCR-epitope prediction exhibits distinct TCR and peptide feature preferences as well as differing trajectories of model certainty.
\end{abstract}
    \section{Introduction}
Scientific applications across biology~\citep{chen2024applying}, medical image analysis~\citep{shen2017deep}, and neuroscience~\citep{richards2019deep} have benefited from the effectiveness of deep learning~\citep{jordan2015machine}. However these models operate as black boxes, making their decision-making mechanisms difficult to understand and limiting their transparency and trustworthiness~\citep{arrieta2020explainable}. Critically, lack of transparency also confounds researchers' ability to gather scientific insights and also to learn and diagnose model failures, and subsequently improve performance~\citep{chen2024applying}. Consequently, explainable artificial intelligence (XAI) has received significant attention in both academia and industry~\citep{dwivedi2023explainable}. For example, Anthropic and OpenAI have investigated XAI to understand the internal mechanisms of AI models~\citep{anthropic2025biology,gao2025weight}, while academia has developed methods for interpreting black-box models in computer vision~\citep{chen2024applying}, natural language processing~\citep{zhao2024explainability}, and protein design~\citep{hunklinger2026towards}.

AI for science problems are particularly challenging because scientific data are often limited and the underlying biological mechanisms remain incompletely understood~\citep{wang2023scientific}. 
This is particularly evident in T cell receptor (TCR)-epitope binding prediction, an important problem in adaptive immunity and a key step toward vaccine design and immunotherapy development~\citep{joglekar2021t,rojas2023personalized,poorebrahim2021tcr}. From a machine learning perspective, TCR-epitope binding prediction is a binary classification problem using up to four inputs: the TCR $\alpha$ chain, TCR $\beta$ chain, peptide, and major histocompatibility complex (MHC) molecule, with each TCR chain containing three complementarity-determining regions (CDRs)~\citep{hudson2023can}. Numerous models have been developed using diverse architectures, including transformers such as TULIP~\citep{meynard2024tulip} and MixTCRpred~\citep{croce2024deep}, and convolutional neural networks (CNNs) such as NetTCR-2.2~\citep{jensen2023nettcr}. However, recent benchmarks, including IMMREP23~\citep{nielsen2024lessons}, IMMREP25~\citep{richardson2026immrep25}, and \cite{lu2026assessment} benchmark, show that their generalization to unseen epitopes remains limited, while their black box nature makes it difficult to determine why and how they fail. Recent studies have addressed this challenge using post-hoc interpretation, such as QCAI for TULIP~\citep{li2025quantifying}, and interpret-by-design approaches, such as TCR-SRIM~\citep{li2026structure} and TCR-EML~\citep{li2025tcr}. These methods provide insights into the evidence used by models and how they classify TCR-epitope pairs, and QCAI-guided model design has further demonstrated that such insights can improve prediction models~\citep{li2025rational}. However, these approaches primarily interpret trained models and do not reveal why particular evidence is learned, ignored, or organized during training~\citep{dwivedi2023explainable}. We therefore adopt a definition of explainability as the organization and evolution of interpretable evidence into a coherent structure, rather than the extraction of evidence from a trained black-box model after inference~\citep{miller2019explanation,lipton2018mythos}. Understanding this learning process is essential for explaining TCR-epitope binding prediction mechanisms. For example, recent work shows that an unpaired TCR chain can be sufficient for prediction~\citep{shah2026unpaired}. Post-hoc interpretability alone cannot explain why do the two chains provide conflicting information, highlighting the need to explain the learning process itself.

Our primary contribution this paper is to develop \ModelName[highlight] (\ModelName[abbr]), a new paradigm that  tracks the evolution of model interpretations throughout training to provide structured model explanations. \ModelName[abbr] is model-agnostic and can be applied to both interpret-by-design models (e.g., TCR-SRIM~\citep{li2026structure}), and black box models with post-hoc interpretations, including transformer-based TULIP~\citep{meynard2024tulip} and CNN-based NetTCR-2.2~\citep{jensen2023nettcr}.
Our second contribution is to develop TCR-XAI2, an extension of a crystal structure-based TCR-XAI benchmark, which contains 274 TCR-epitope structures~\citep{li2025quantifying}; we extend TCR-XAI it by incorporating 144 additional experimentally resolved structures from  TCR3D2.0~\citep{lin2025tcr3d}. TCR-XAI2 also includes structure predictions from AlphaFold3~\citep{abramson2024accurate}, tFold-TCR~\citep{wu2025fast}, Boltz-2~\citep{passaro2025boltz}, OpenFold3~\citep{openfold3-preview}, and TCRModel2~\citep{yin2023tcrmodel2}, to allow a examination of the impact of experimental versus predicted structures in any particular TCR-epiope model that utilizes structure. We also annotate CDR and peptide residues according to their structural roles.

Our third contribution is the application of \ModelName[abbr] and TCR-XAI2 to TCR-SRIM, TULIP, MixTCRpred, and NetTCR-2.2 to analyze their learning trajectories and characterize differences in how they learn TCR-epitope binding. Our studies reveal three findings: (1) CNN-based models rely more heavily on peptide information, whereas transformer-based models place greater emphasis on TCR information; (2) models without MHC input consistently exhibit opposing changes in interpretation quality between the TCR $\alpha$ and $\beta$ chains, while MHC information reduces this effect, suggesting that models may struggle to effectively integrate paired TCR chains;
and (3) there are significant differences in TCR and peptide feature preference and model certainty during training when using real versus predicted structures for model regularization.

    \section{Our Approach}
\begin{figure}[t]
    \centering
    \includegraphics[width=\linewidth]{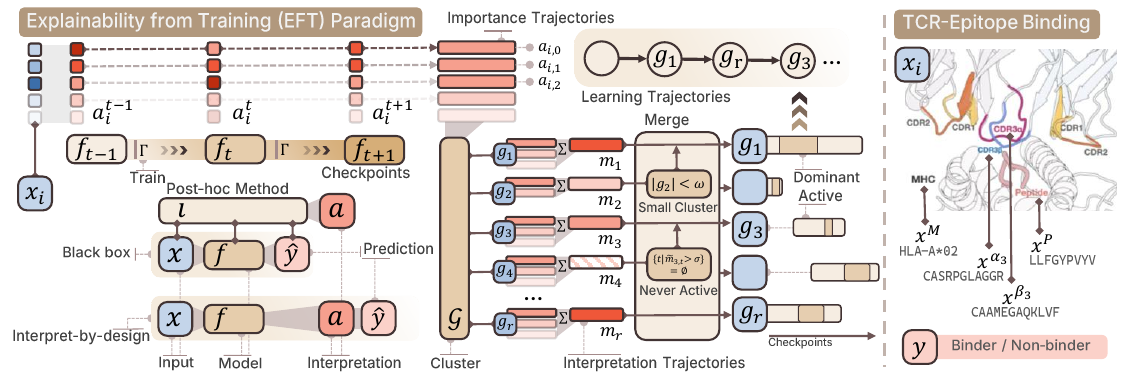}
    \caption{The \ModelName[abbr] paradigm tracks model interpretations during training to obtain importance trajectories, clusters them into interpretation trajectories, and organizes them into learning trajectories that provide explanations. The diagram of TCR-epitope binding is from ~\cite{parizi2026swifttcr}.}
    \label{fig:pipeline}
\end{figure}

We introduce our \ModelName[norm] (\ModelName[abbr]) paradigm based on unified definitions of binary classification models and their corresponding interpretation methods. As shown in Fig.~\ref{fig:pipeline}, \ModelName[abbr] tracks changes in model interpretations throughout training to characterize their learning trajectories. We first track feature importance changes to obtain importance trajectories, then cluster features according to the similarity of their trajectories to obtain cluster-level interpretation trajectories. Finally, we organize these trajectories into model learning trajectories, providing explanations of why a model relies on specific features and how it organizes them as predictive evidences during training.

\paragraph{Predictive Models \& Corresponding Interpretations.}
We first define binary classification models and their corresponding interpretations. Given an input $x\in\mathcal{X}=\mathbb{R}^N$, that is a $N$ elements real number vector, a binary classification model produces a probability prediction $\hat{y}\in\hat{\mathcal{Y}}=[0,1]$, and can be represented as $f:\mathcal{X}\rightarrow\hat{\mathcal{Y}},\, f\in\mathcal{F}.$
A dataset is defined as
\begin{equation}
D={(x_i,y_i)}_{i=1}^{N_{\mathrm{D}}},
\qquad x_i\in\mathcal{X},\quad y_i\in\mathcal{Y}=\{0,1\},
\end{equation}
where $N_{\mathrm{D}}$ denotes the number of input-label pairs. We define a training operator $\mathcal{T}:\mathcal{F}\times D\rightarrow\mathcal{F},$ which updates a model checkpoint using the training dataset. Starting from checkpoint $f_t$, the checkpoint after one training step is given by $f_{t+1}=\mathcal{T}(f_t,D).$

For a given checkpoint $f_t$, input $x$, and prediction $\hat{y}=f_t(x)$, its interpretation can be obtained in two ways. For black-box models, we use a post-hoc method $\iota$ to obtain feature importance scores:
\begin{equation}
\iota:\mathcal{F}\times\mathcal{X}\times\hat{\mathcal{Y}} \rightarrow\mathcal{A}, \qquad \mathcal{A}=[0,1]^{N},
\end{equation}
where $\mathcal{A}$ denotes the space of normalized feature importance scores with the same dimensionality as $x$. Different interpretation methods may produce different interaction structures. For example, GradCAM and AttnLRP assign a single importance score to each input feature, whereas QCAI can produce interaction-specific scores for features participating in different modalities. When multiple importance scores are assigned to the same feature, we average them to obtain a unified importance score.
For interpret-by-design models, intermediate evidence is explicitly constructed as part of the prediction process. We therefore directly extract the corresponding feature importance scores from this evidence and transform them into the same representation $\mathcal{A}$.

\paragraph{Tracking Interpretations During Training.}
We first select a subset of samples as the tracking set, denoted by $D^{\mathrm{T}}\subseteq D$.
Given the unified definitions of models and interpretation methods, we obtain the prediction of each checkpoint $f_t$ for every sample in the tracking set as $f_t(x_i)=\hat{y}_i^t, (x_i,y_i)\in D^{\mathrm{T}}.$
We then apply the corresponding interpretation method $\iota$ to obtain the importance
\begin{equation}
\iota(f_t,x_i,\hat{y}_i^t)=a_i^t\in\mathcal{A}^t, \, (x_i,y_i)\in D^{\mathrm{T}},
\end{equation}
where $\mathcal{A}^t$ is all importance scores of $f_t$ on $D^\mathrm{T}$.
By repeating this process across all checkpoints, we obtain a trajectory of importance $\{\mathcal{A}^t\}_{t=0}^{N_T}$ for each tracking sample, which characterizes how the model's interpretation evolves throughout training.

\paragraph{Building Importance Trajectories.}
For each sample $x_i$ and $f_t$, let $a_i^{t}$ denote the corresponding vector of feature importance scores. For an input $x_i$, we represent the importance scores as
\begin{equation}
a_i^{t}=\left[a_{i,j}^{t}\right]_{j=0}^{N}=
\left[a_{i,0}^{t},a_{i,1}^{t},\ldots,a_{i,N-1}^{t}\right],
\end{equation}
where $a_{i,j}^{t}\in[0,1]$ denotes the importance of position $j$ in input $x_i$ at checkpoint $t$. For a fixed position $j$, its importance throughout training can therefore be represented as the importance trajectory
\begin{equation}
\mathbf{a}_{i,j} = \left[a_{i,j}^{t}\right]_{t=0}^{N_T} =
\left[a_{i,j}^{0},a_{i,j}^{1},\ldots,a_{i,j}^{N_T}\right] \in[0,1]^{N_T}.
\end{equation}

To identify when a feature becomes important during training, we define an activation threshold based on its importance trajectory. Given a $\tau\in(0,1]$ representing the top-$\tau$ fraction of checkpoints, we consider feature $x_{i,j}$ to be active at checkpoint $t$ if its importance is within the top $\tau$ fraction of its importance values throughout training. The set of active checkpoints is therefore defined as
\begin{equation}
\pi_{i,j} = \left\{ t\;\middle|\; a_{i,j}^{t}\geq q_{1-\tau}\left(\mathbf{a}_{i,j}\right), \, t\in\{0,\ldots,N_T\}\right\},
\end{equation}
where $q_{1-\tau}:\mathbb{R}^{N_T}\rightarrow\mathbb{R}$ denotes the quantile function. Thus, $\pi_{i,j}$ identifies the checkpoints at which a feature exhibits relatively high importance compared with its own importance throughout training.

\paragraph{Feature Clustering and Interpretation Trajectories.}
Analyzing importance trajectories of individual features produce a large number of trajectories and can substantially reduce the readability of the resulting explanations. We therefore cluster features according to the consistency of importance trajectories throughout training to obtain interpretation trajectories. To make importance trajectories comparable across samples and positions, we normalize each of them across checkpoints:
$\tilde{\mathbf{a}}_{i,j} = \mathbf{a}_{i,j} / \sum\nolimits_{t=0}^{N_T}a_{i,j}^{t}.$
We then apply K-means clustering to the normalized importance trajectories: $\mathcal{G}_K\left(\tilde{\mathbf{a}}\right) = \{g_r\}_{r=1}^{K},$
where $\mathcal{G}_K$ denotes K-means clustering with $K$ clusters (interpretation trajectories). Each cluster $g_r$ contains indices $(i,j)$ corresponding to a specific sample and position.
For each cluster, we summarize its importance at each checkpoint using the active features. Specifically, we define
\begin{equation}
m_{r,t} =  \left( \sum\nolimits_{(i,j)\in g_{r,t}} a_{i,j}^{t} \right ) / |g_{r,t}|,
\qquad
g_{r,t} = \left\{(i,j)\in g_r \mid t\in\pi_{i,j} \right\},
\end{equation}
where $g_{r,t}$ denotes the subset of features in cluster $g_r$ that are active at checkpoint $t$. We then normalize the cluster importance across training checkpoints: $\tilde{m}_{r,t} = m_{r,t} / \max\nolimits_{t'}m_{r,t'}.$
A cluster is considered active at checkpoint $t$ when $\tilde{m}_{r,t}>\sigma, \sigma\in(0,1].$
In addition to identifying active clusters, we seek to determine which evidence dominates the model's learning process at each checkpoint. We define cluster $g_r$ as dominant at checkpoint $t$ if it has the highest cluster importance and its importance exceeds a threshold $\varepsilon$:
\begin{equation}
\mathcal{U}_r^{\mathrm{d}} = \left\{ t\;\mid\; r=\arg\max\nolimits_{r'} m_{r',t}, \quad \max\nolimits_{r'}m_{r',t}>\varepsilon \right\}.
\end{equation}

Finally, we merge clusters that provide limited explanatory value to reduce the complexity of the resulting learning trajectories, also training-time explanations. A cluster $g_r$ will be merged to nearest cluster $g_m$, if it is small: $|g_r|<\omega$, where $\omega\in\mathbb{Z}^{+}$, or if it is never active during training: $\{t\mid \tilde{m}_{r,t}>\sigma,\ t\in[0,N_{\mathrm{T}}]\}=\emptyset.$
Then, this cluster $g_r$ is merged into the nearest remaining cluster $g_{m}$, where
\begin{equation}
m = \arg\min\nolimits_{r'\in[1,K]\setminus{r}} \sum\nolimits_{t=0}^{N_{\mathrm{T}}}
\left| \left(\sum\nolimits_{(i,j)\in g_r}\tilde{a}_{i,j}^{t}\right)/|g_r|
-
\left(\sum\nolimits_{(i,j)\in g_{r'}}\tilde{a}_{i,j}^{t}\right)/|g_{r'}|
\right|.
\end{equation}


\subsection{Applying EFT to TCR-epitope Prediction}
TCR-epitope binding prediction can be formulated as a binary classification problem. We consider two types of inputs: amino acid sequences and MHC alleles. Amino acid sequences are represented as strings over the amino acid alphabet $\Theta$ (e.g., \texttt{CASSIVGGNEQFF}), while MHC alleles are represented as the allele set $\mathcal{M}$ (e.g., \texttt{HLA-A*02}). A TCR consists of an alpha and a beta chain, each containing three CDRs, which can be denoted as $x^{\mathrm{T}} = \left(x^c\right)_{c \in \mathcal{C}},\,\mathcal{C} = \{\alpha_1,\alpha_2,\alpha_3,\beta_1,\beta_2,\beta_3\}, $
where $x^c \in \Theta^{N_{\mathrm{C}}}$ denotes the amino acid sequence of CDR region $i$. For simplicity, we assume that all CDRs are padded to a common length $N_{\mathrm{C}}$. For example, CDR3a and CDR3b are represented by $x^{\alpha_3}$ and $x^{\beta_3}$, respectively. The peptide and MHC allele are represented as $x^{\mathrm{P}} \in \Theta^{N_{\mathrm{P}}}$ and $x^{\mathrm{M}} \in \mathcal{M}$, respectively, where $N_{\mathrm{P}}$ denotes the peptide length.
A TCR-epitope sample is therefore represented as $x = x^{\mathrm{T}}\cup x^{\mathrm{P}}\cup x^{\mathrm{M}} \in \mathcal{X},$
where different models may use all or a subset of these inputs.
Black box TCR-epitope models can be interpreted using standard post-hoc methods, whereas interpret-by-design models explicitly construct evidence for prediction. For example, TCR-SRIM produces contact prototypes that represent interactions between two input sequences before generating the final prediction. A contact prototype is represented as an interaction matrix $\phi\in\mathbb{R}^{N_{\mathrm{C}}\times N_{\mathrm{P}}}$, where $\phi^{\alpha\rightarrow\mathrm{P}}$ and $\phi^{\beta\rightarrow\mathrm{P}}$ denote the prototypes for CDR3a-peptide and CDR3b-peptide interactions, respectively. To obtain residue-level importance scores in the unified representation, we average each contact prototype along the corresponding dimension.

    \section{Results and Discussion}
In this section,
we first describe applying \ModelName[abbr] to trace the training evolution of explanations for four TCR-epitope prediction models: NetTCR-2.2~\citep{jensen2023nettcr}, MixTCRpred~\citep{croce2024deep}, TULIP~\citep{meynard2024tulip}, and TCR-SRIM~\citep{li2026structure}.
NetTCR-2.2 is a CNN model, while the rest of models are transformer-based. Both NetTCR-2.2 and MixTCRpred concatenate all CDR regions as input, meanwhile, TULIP and TCR-SRIM using cross-attention to fuse only CDR3 and peptide inputs. TULIP provides two versions of model that with or with our MHC allele as input. TCR-SRIM is the only interpret-by-design model that models CDR3-peptide interactions are contact maps, which allows to be trained with or without structure regularization. We first compare the version without structure regularization to align configurations of other models. Second, we discuss structure information in training using TCR-SRIM with structure regularization to compare the affects of experimental vs. predicted structures.
We use the optimal interpretation method for each model: GradCAM~\citep{selvaraju2017grad} for NetTCR-2.2, AttnLRP~\citep{achtibatattnlrp} for MixTCRpred, and QCAI~\citep{li2025quantifying} for TULIP. Each experiment is repeated three times with random seeds 0, 1, and 2. Ablation studies on interpretation methods and random seeds are provided in Appendix~\ref{sec:diffinterpmethod} and Appendix~\ref{sec:diffrunseed}, respectively.


Next, we extend TCR-XAI benchmark~\citep{li2025quantifying} to build TCR-XAI2 dataset, provide an overview of its composition, and compare it with TCR-XAI.
We quantitatively analyze these evolving explanations to investigate the potential conflicts between TCR $\alpha$ and $\beta$ chains, the role of MHC alleles in model learning, and the TCR-epitope unbinding mechanism from the model's perspective. We further present case studies to illustrate these observations on specific examples. Finally, we compare TCR-SRIM models regularized with structures predicted by different structure prediction models to examine how structural information influences model learning. Based on these observations, we discuss implications for designing improved TCR-epitope prediction models.

\subsection{TCR-XAI2 Benchmark}
\cite{li2025quantifying} introduced TCR-XAI, a benchmark of 274 experimentally resolved TCR-epitope structures with residue-level contact maps as ground-truth explanations for quantitatively evaluating interpretation quality. To expand the benchmark, we collected the latest structures from the continuously updated TCR3D2.0~\citep{lin2025tcr3d} and STCRDab~\citep{leem2018stcrdab}, yielding 388 unique PDB structures. Since a single PDB entry may contain multiple copies of a TCR-epitope complex, we treated each copy as an independent sample, resulting in 619 samples. We further predicted the structures of these samples using AlphaFold3~\citep{abramson2024accurate}, Boltz-2~\citep{passaro2025boltz}, TCRModel2~\citep{yin2023tcrmodel2}, OpenFold3~\citep{openfold3-preview}, and tFold-TCR~\citep{wu2025fast}. All methods generated predictions for the full set of samples except AlphaFold3 and TCRModel2, which failed for 15 and 28 samples, respectively. TCRModel2 additionally models only the TCR variable domains and peptide-binding groove, excluding $\beta_2$-microglobulin, the TCR constant domains, and the MHC $\alpha_3$ domain. We combine these experimentally resolved and predicted structures to construct the TCR-XAI2 benchmark.
We also provide training and test splits of TCR-XAI2 for models that require structural information during training, such as TCR-SRIM. To reduce sequence and structural leakage between the two splits, we evaluated the minimum Levenshtein distance between CDR3 and peptide sequences across the training and test sets. The median minimum distance is 18 residues, with a minimum of 5 residues, indicating substantial sequence differences between the splits. We further assessed structural differences using CDR3a-peptide and CDR3b-peptide contact maps, obtaining a median average contact-map distance of 2.65~\AA.

\paragraph{Residue-level Structure Labels.}
We assign structural annotations to each residue in the CDR3a, CDR3b, and peptide regions of the TCR-XAI2 benchmark. These annotations describe each residue's structural role at the binding interface and are computed from either experimentally resolved or predicted structures. A contact label (e.g., \texttt{contact CDR3b}) is assigned when the closest heavy atom of a residue is within $5\,\text{\AA}$ of a residue in the corresponding partner region. A burial label (e.g., \texttt{buried CDR3b}) identifies a residue that buries more than $1\,\text{\AA}^2$ of surface area from a partner upon binding. We calculate this value as the reduction in solvent-accessible surface area using the Lee-Richards algorithm with a probe radius of $1.40\,\text{\AA}$.
An geometric annotation (e.g., \texttt{geom+ CDR3b}) captures geometric rather than distance-based interactions and identifies residues that define the docking frame of the complex. Specifically, we use the conserved IMGT anchor positions $104$ and $118$ on each CDR3 loop and the N- and C-terminal residues of the peptide~\citep{sanou2026imgt}.  A residue may receive multiple labels because different structural roles can occur simultaneously, such as when a peptide residue contacts multiple CDR3 loops.


\subsection{Model Explanations from Training}
\begin{figure*}[t]
    \centering
    \includegraphics[width=0.49\linewidth]{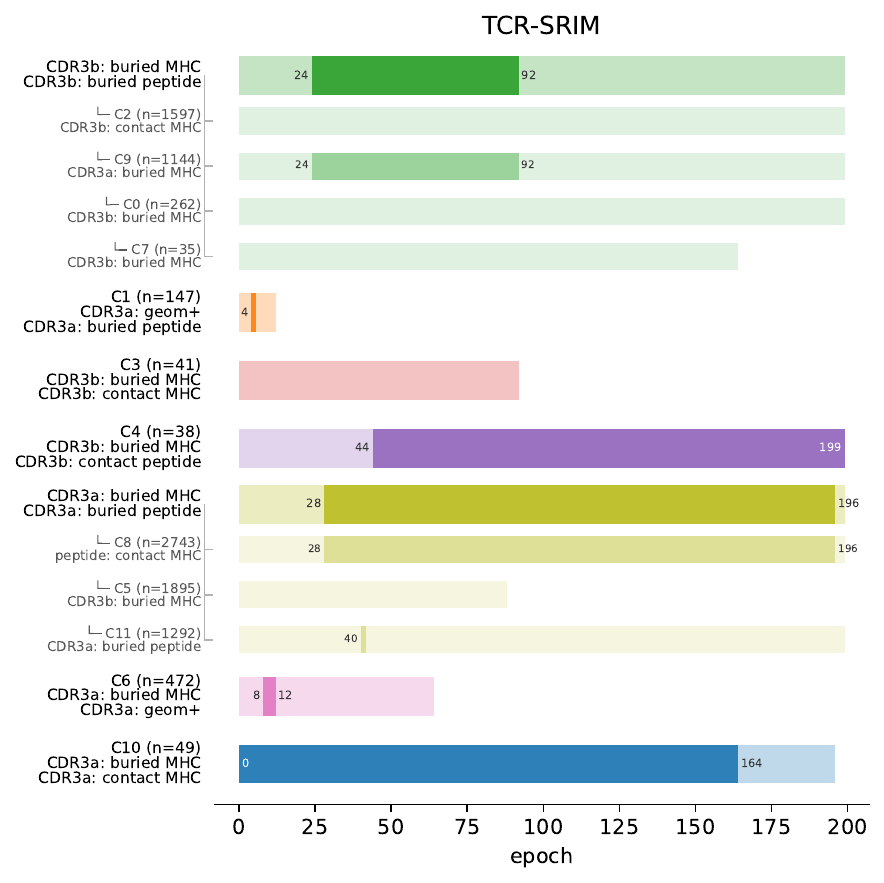}
    \includegraphics[width=0.49\linewidth]{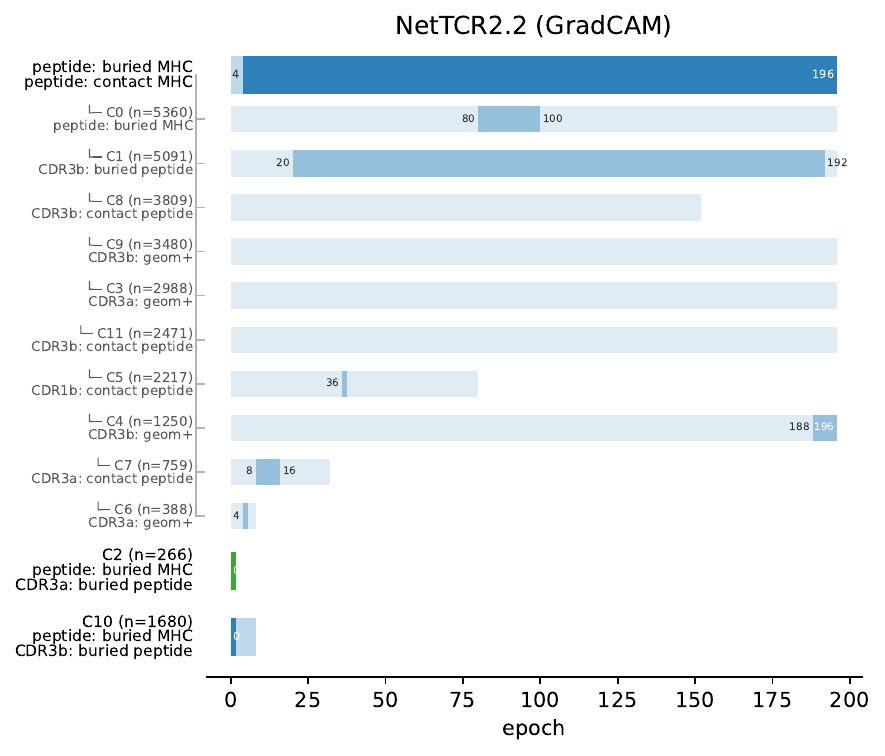}
    \includegraphics[width=0.49\linewidth]{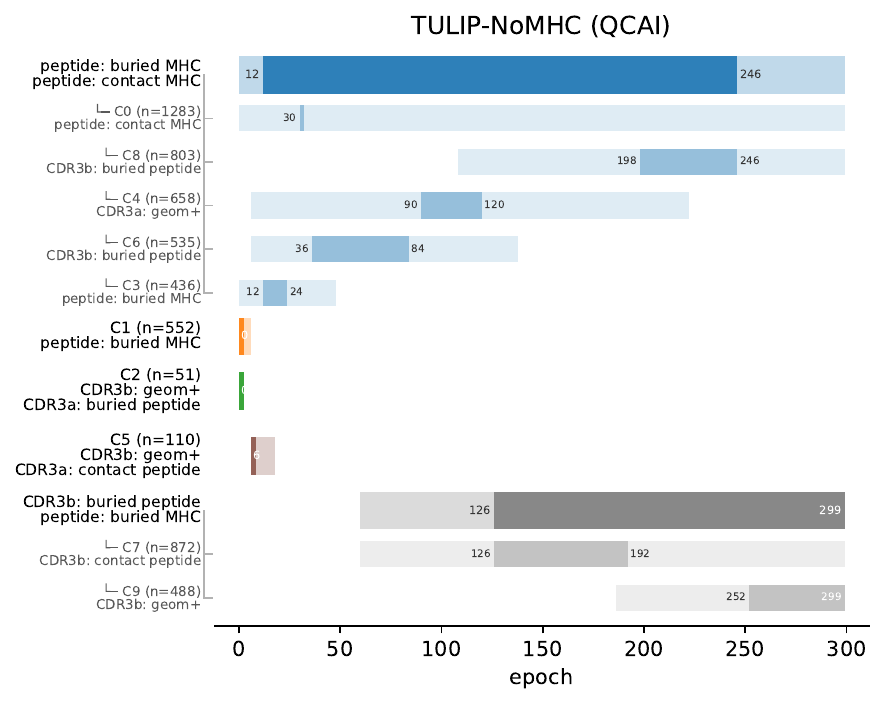}
    \includegraphics[width=0.49\linewidth]{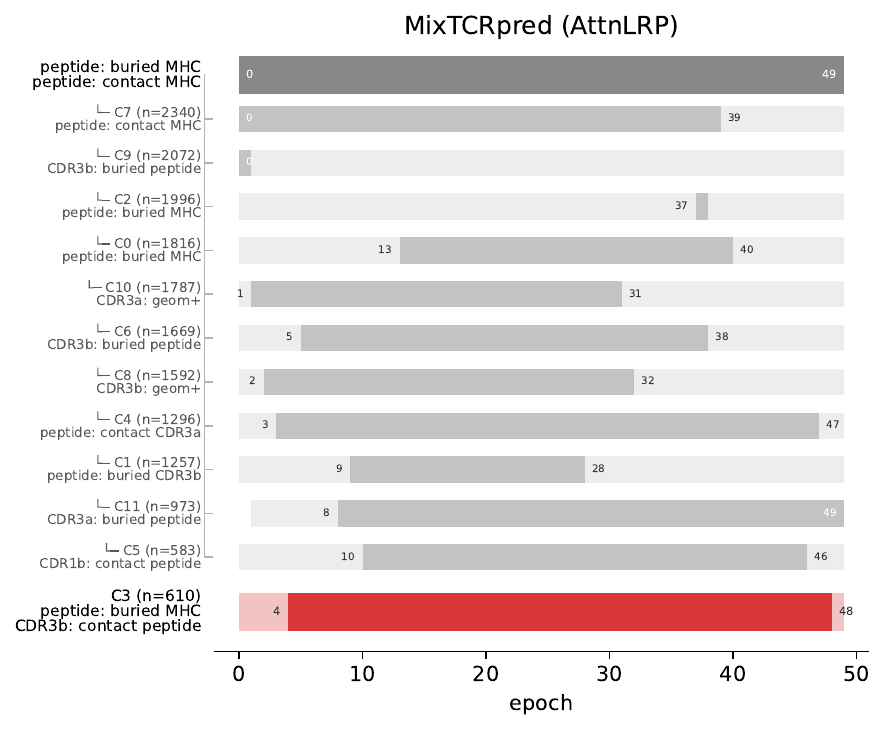}
    \caption{\ModelName[abbr] explanations reveal distinct learning trajectories across CNN and transformer and feature fusion strategies, while showing that models struggle to learn TCR $\alpha$ and $\beta$ simultaneously.
}
    \label{fig:modelexplain:general}
    \vspace{-0.5cm}
\end{figure*}

Using \ModelName[abbr], we first analyze the training-based explanations of four TCR-epitope binding prediction models: TCR-SRIM without structural regularization, TULIP without MHC allele as input, NetTCR-2.2, and MixTCRpred with peptide information as input. NetTCR-2.2 and MixTCRpred use all CDR regions as model inputs.
As shown in Fig.~\ref{fig:modelexplain:general}, features are grouped into clusters according to their model representations. The light bars indicate features that are active during a given training period, while darker bars indicate clusters that dominate the model's explanation during that period.
We use the structural labels defined from experimentally resolved structures to characterize each feature cluster. Because approximately 75\% of peptide residues are buried within the MHC binding groove, the labels \texttt{peptide buried MHC} and \texttt{peptide contact MHC} dominate the training process of most models. TCR-SRIM is an exception because its contact prototype layers are explicitly designed to capture TCR-peptide interactions. To provide more informative descriptions, we report the top two or three structural labels for each cluster. We first characterize each cluster using its top two labels and group clusters with the same top two labels into a super-cluster. We then use the third-ranked label to distinguish and characterize the child clusters within each super-cluster.

The models primarily rely on CDR3 regions and peptides for TCR-epitope binding prediction. Even for NetTCR-2.2 and MixTCRpred, only one feature cluster is associated with CDR1b-peptide contact. Although both models use the same inputs, NetTCR-2.2 primarily focuses on CDR3b burial of the peptide, which dominates across most training epochs, whereas MixTCRpred emphasizes CDR3a and exhibits more diverse dominant feature clusters. These differences suggest that the CNN-based NetTCR-2.2 and transformer-based MixTCRpred organize TCR-epitope evidence differently, with MixTCRpred distributing evidence across a broader range of interaction features.

Beyond these observations, we identify two additional patterns in the training process. For models other than TCR-SRIM, the models consistently rely on peptide-MHC interactions throughout training, particularly during the early and late stages. In contrast, CDR-peptide and CDR-MHC interactions primarily emerge during intermediate stages of training. Furthermore, the models tend to learn CDR3a-related information before CDR3b-related information, and do not learn TCR alpha and beta information simultaneously.


\begin{figure}[t]
    \centering
    \includegraphics[width=\linewidth]{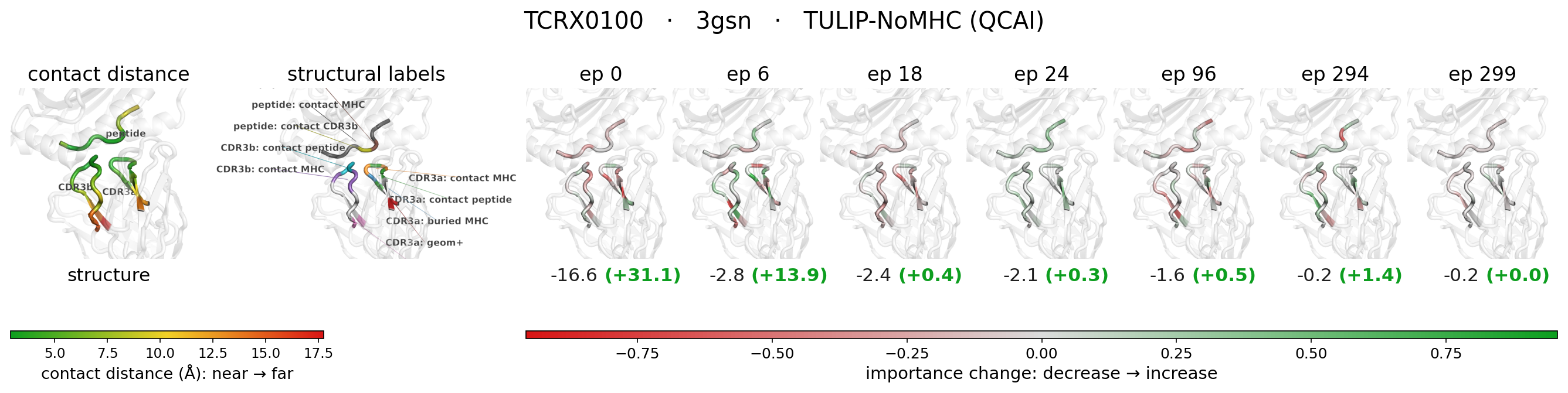}
    \caption{A \ModelName[abbr] case study of TULIP-NoMHC on a human cytomegalovirus sample reveals conflicts between TCR $\alpha$ and $\beta$ information for model learning during training.
    }
    \label{fig:casestudy}
\end{figure}

We then present a case study using \ModelName[abbr] to analyze TULIP without MHC allele information to demonstrate \ModelName[abbr] explanation for a single case. It corresponds to a human cytomegalovirus infection sample from PDB structure \texttt{3GSN}. Consistent with the general explanation analysis, TULIP initially focuses on the CDR3a-peptide interaction and progressively increases its attention to peptide residues involved in CDR3a contacts during the early training stages.
We observe a clear contrast between the alpha and beta chains from epochs 0 to 18. The residue-level attention patterns shift in opposite directions between the two chains, indicating an alpha-beta conflict in the evidence used by the model. This can explain why model does not learn TCR $\alpha$ and $\beta$ information simultaneously.


\subsubsection{Role of MHC and TCR Chain Conflicts in Model Learning}
\begin{figure*}[t]
    \centering
    \includegraphics[width=0.49\linewidth]{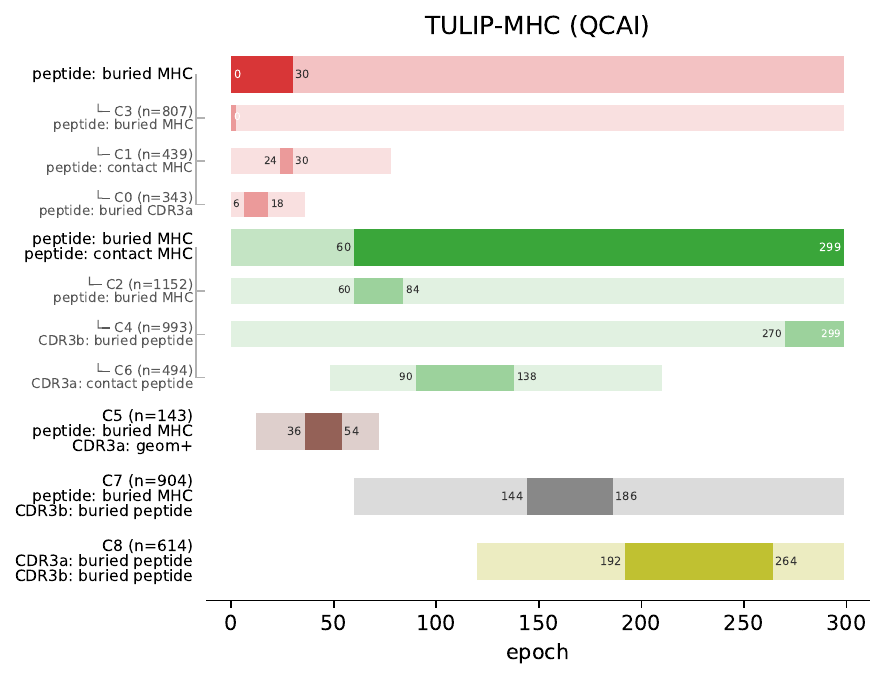}
    \includegraphics[width=0.49\linewidth]{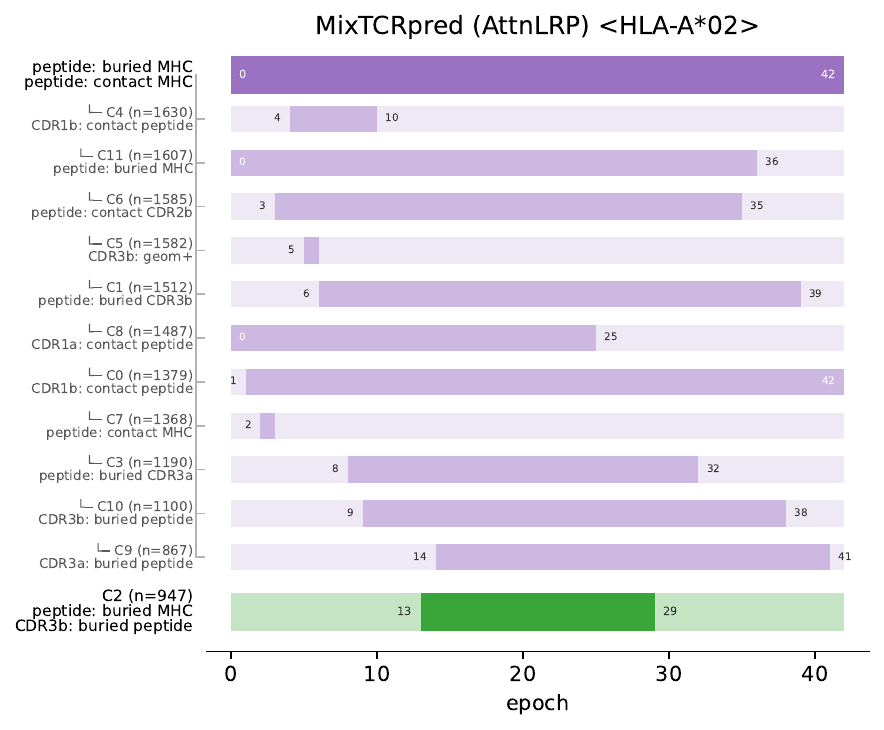}
    \caption{\ModelName[abbr] explanations for TULIP-MHC and MixTCRpred trained on samples with only the \texttt{HLA-A*02} allele show that MHC information can reduce conflicts between TCR $\alpha$ and $\beta$ information and enable models to learn features from more chains.
    }
    \label{fig:modelexplain:mhc}
\end{figure*}
To further investigate why models cannot learn TCR alpha and beta information simultaneously and how MHC alleles can influence TCR-epitope model learning, we analyze \ModelName[abbr] explanations for TULIP with MHC allele information as an input and for MixTCRpred trained exclusively on samples with the \texttt{HLA-A*02} allele. As shown in Fig.~\ref{fig:modelexplain:mhc}, providing MHC allele information or restricting training data to a specific MHC allele leads to distinct learning trajectories. For TULIP with MHC information, a feature cluster corresponding to CDR3a and CDR3b burial of the peptide remains dominant for more than 50 epochs starting from epoch 192. In contrast, for the model without MHC information, the CDR3a-peptide and CDR3b-peptide interaction features never become dominant simultaneously. 
The difference is more pronounced for MixTCRpred trained exclusively on \texttt{HLA-A*02} samples, where CDR1a and CDR2b also emerge as important evidence. These observations suggest that MHC information can play two complementary roles in TCR-epitope model learning. When MHC information is not provided and the training data cover diverse MHC alleles, the model may need to infer MHC-dependent information indirectly from the TCR and peptide sequences. This can lead the model to associate peptide recognition with one TCR chain as a proxy for MHC-dependent effects. In contrast, when MHC information is provided or the training data are restricted to a specific MHC allele, the model can allocate more capacity to other input features and jointly integrate information from the alpha and beta chains.

In other hand, this learning process may create a conflict between the two TCR chains when MHC information is unavailable. Before learning to coordinate both chains, the model may rely more heavily on one chain to recognize the peptide, limiting its use of complementary information from the other chain. This provides a potential explanation for why incorporating paired TCR chains does not necessarily yield substantially higher predictive performance than using unpaired chains~\citep{shah2026unpaired}. This is also the reason why \cite{li2025rational} finds adding a cross-attention to fusion TCR alpha and beta chains information can improve both interpretation quality and performance.

\subsection{Interpretability During Training}
To further examine the roles of MHC information and potential conflicts between the alpha and beta chains during model learning, we quantitatively track binding region hit rate (BRHR) throughout training for the peptide, CDR3a, and CDR3b regions, which is introduced by ~\cite{li2025quantifying}. We compare TULIP trained with and without MHC allele information. For CDR3a and CDR3b, we measure BRHR@.25 for the CDR3a-peptide and CDR3b-peptide interactions, respectively. For the peptide, we use the average BRHR@.25 across its interactions with CDR3a and CDR3b.

\begin{figure}[t]
    \centering
    \includegraphics[width=0.48\linewidth]{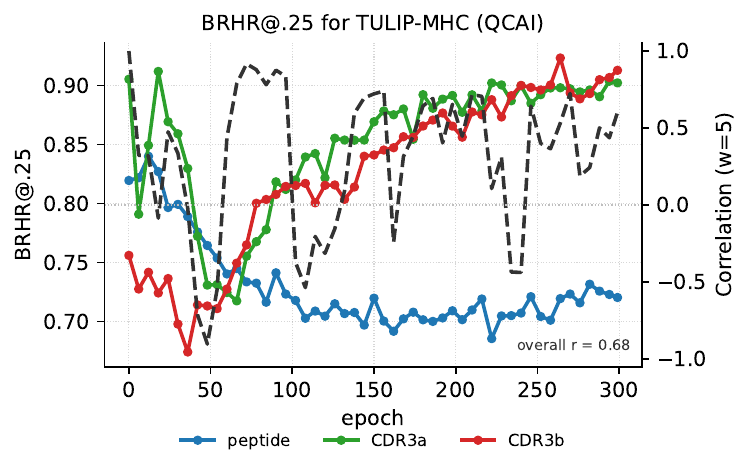}
    \includegraphics[width=0.48\linewidth]{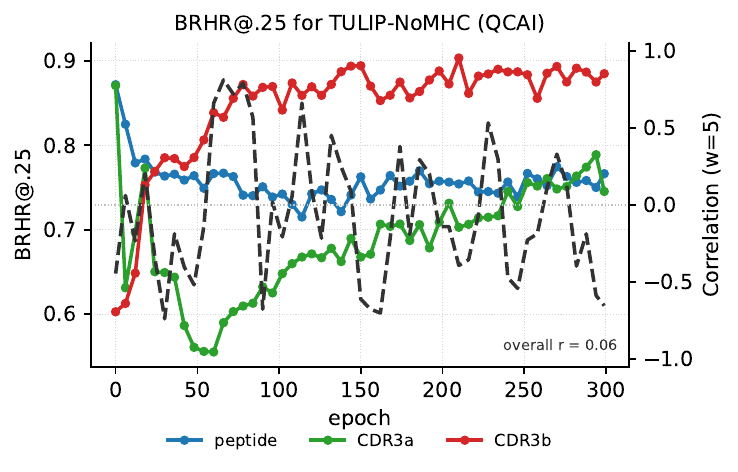}
    \caption{The evolution of BRHR@.25 for CDR3a, CDR3b, and peptide during TULIP training with and without MHC allele information quantitatively demonstrates that MHC information can reduce conflicts between TCR $\alpha$ and $\beta$ information.
    }
    \label{fig:brhrdelta:mixtcrpredattnlrp}
\end{figure}

As shown in Fig.~\ref{fig:brhrdelta:mixtcrpredattnlrp}, the evolution of BRHR during training further supports our previous observations regarding the role of MHC information and potential conflicts between the TCR alpha and beta chains. In addition to tracking the BRHR of CDR3a and CDR3b separately, we compute their sliding-window Pearson correlation using a window size of 5. When MHC allele information is provided as an input, the BRHR trajectories of CDR3a and CDR3b are substantially more consistent, with an overall Pearson correlation of $r=0.68$. In contrast, without MHC information, the BRHR difference between CDR3a and CDR3b exceeds 0.1, while their overall correlation is only $r\approx0.06$. Notably, strong fluctuations and opposing trends are observed during the early stages of training.
These results further support our previous observation that providing MHC allele information reduce the effort required to infer the peptide-MHC relationship indirectly, allowing the model to allocate more capacity to coordinating information from the alpha and beta chains. The consistent BRHR trajectories observed when MHC information is available suggest that the model can learn more coordinated evidence from the two TCR chains.

\subsection{Experimental versus Predicted Structure for Guiding Model Training}

\begin{figure}[t]
    \centering
    \includegraphics[width=0.48\linewidth]{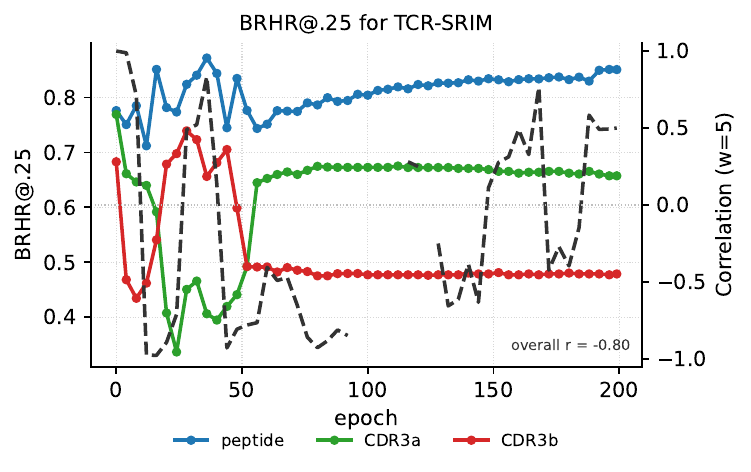}
    \includegraphics[width=0.48\linewidth]{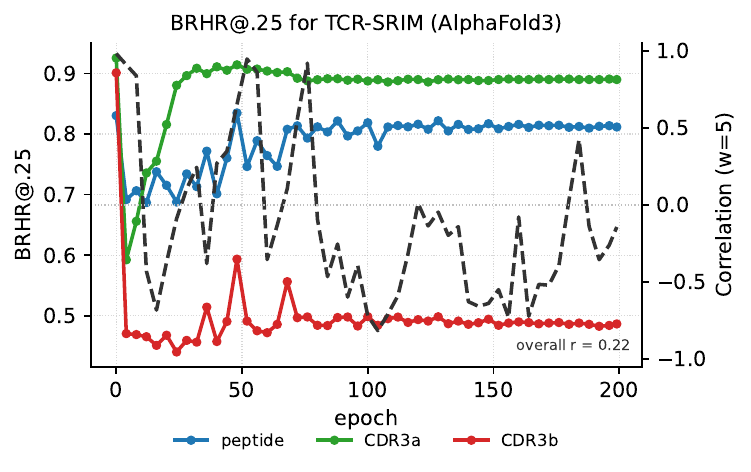}
    \includegraphics[width=0.48\linewidth]{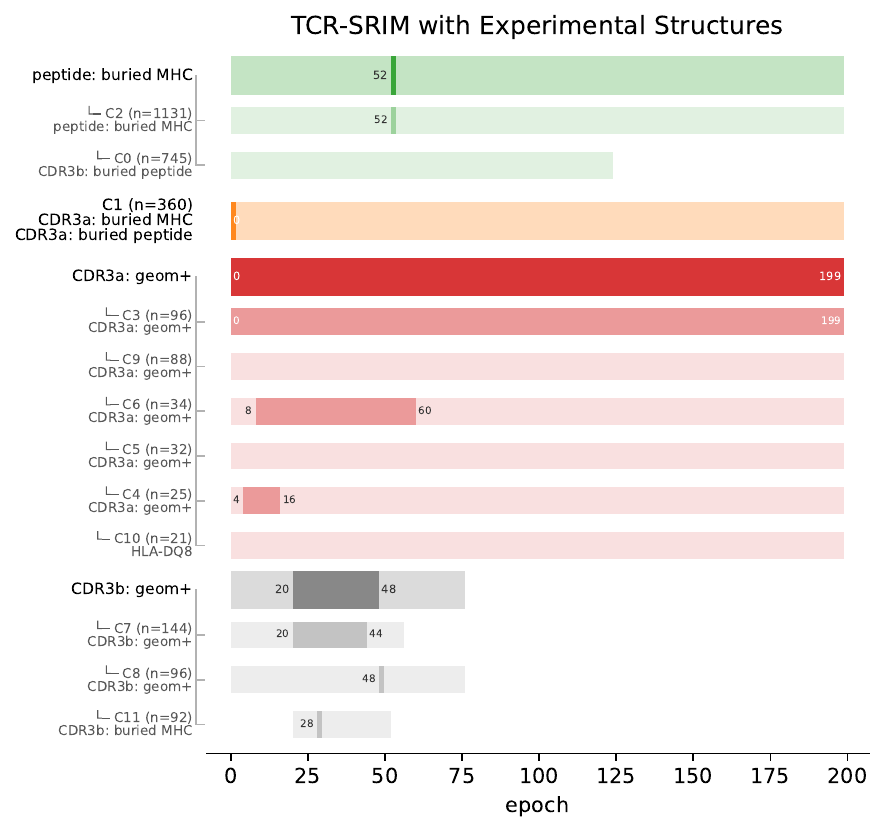}
    \includegraphics[width=0.48\linewidth]{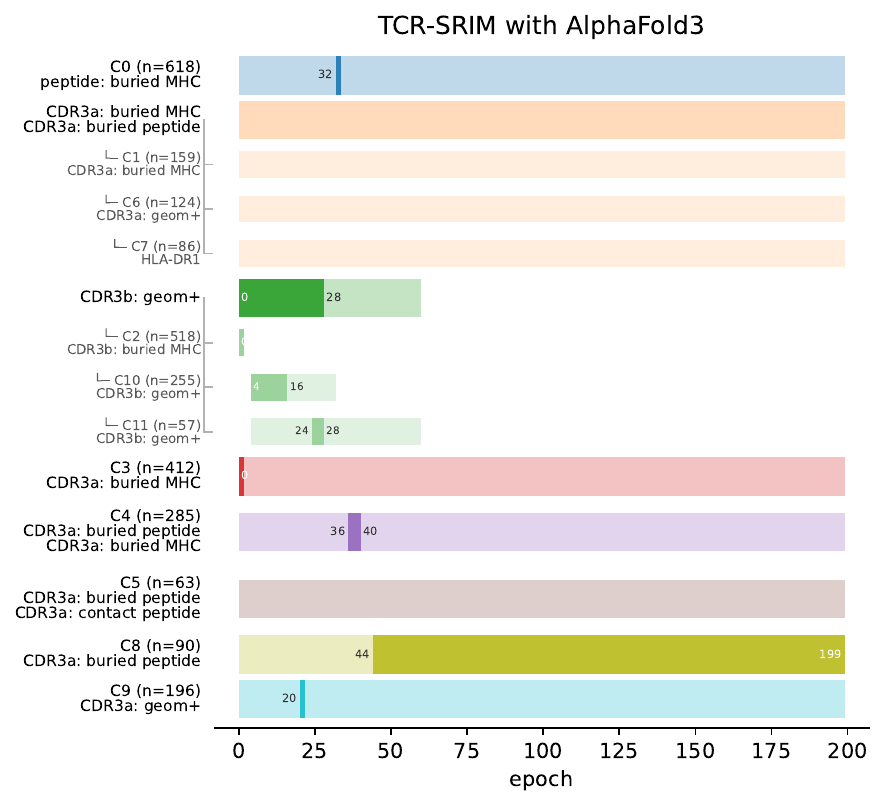}
    \caption{\ModelName[abbr] explanations and evolution of BRHR@.25 during training for TCR-SRIM regularized with experimentally resolved structures and AlphaFold3-predicted structures. Structural information rapidly stabilizes model interpretations, while experimentally resolved and predicted structures lead to different interpretation trajectories during training.
    }
    \label{fig:srim:brhr}
\end{figure}

Recent studies have explored incorporating structural information into TCR-epitope prediction models~\citep{deleuran2025nettcr,li2026structure}. Motivated by these approaches, we use TCR-XAI2 structures predicted by different structure prediction models to investigate how structural quality affects model learning. As a baseline, we first examine \ModelName[abbr] explanations and BRHR evolution in TCR-SRIM without structural regularization. The BRHR trajectories of all three chains fluctuate substantially throughout training rather than converging to stable patterns, indicating that structural evidence remains unstable without structural guidance.
We then regularize TCR-SRIM using experimentally resolved or AlphaFold3-predicted structures from the TCR-XAI2 training split and evaluate BRHR on the test split using experimentally resolved structures. As shown in Fig.~\ref{fig:srim:brhr}, experimentally resolved structures lead to stable BRHR values for all three chains within approximately 70 epochs, with substantially reduced fluctuations. AlphaFold3-predicted structures also stabilize model interpretations. However, \ModelName[abbr] shows that both models primarily focus on CDR3a and CDR3a-peptide interactions, while CDR3b-peptide interactions receive greater attention mainly during early training, consistent with the BRHR analysis. This behavior result from the absence of MHC information, leading the model to rely more heavily on a single TCR chain for prediction.


    \section{Conclusion}
In this paper, we proposed the \ModelName[highlight] (\ModelName[abbr]) paradigm. For a given model and interpretability method, the output of \ModelName[abbr] is a trajectory that captures tracks how the model learns and uses features. To demonstrate its utility, we apply \ModelName[abbr] to four state-of-the-art TCR-epitope models. With the updated TCR-XAI2 benchmark, we analyzed these models and demonstrate that: (1) CNNs rely more on peptide information, while transformers emphasize TCR features; (2) models without MHC input show conflicting TCR $\alpha$- and $\beta$-chain evidence, which is reduced by MHC information; and (3) structural differences between experimentally resolved and predicted structures affect models' preference for TCR or peptide information and trajectories of model certainty.

    \subsection*{AI use statement}


In this work, we used generative AI tools for
implement methods.
We have not used generative AI tools for
Generate synthetic data sets, help develop theoretical models or conceptual frameworks, formulate mathematical claims, provide critical ingredients for proving mathematical claims, assist in the writing of proofs, propose or refine hypotheses, design or provide feedback on research  methodology or experiments, clean and reformat dataset, support qualitative and thematic data analysis, interpret results, 
and
assist with translation
are not applicable to this work.
Additionally, we used generative AI tools for
modify scientific figures or images, create or edit software code,  sourcing/searching for information, edit a research paper to improve readability, propose a title or keywords for a research paper.
We have reviewed all AI-assisted work.
LLM-generated code was verified and tested for correctness.
We take responsibility for the final content of this work,
including text, claims or artifacts produced with the aid of generative AI.
    \subsection*{Reproducibility statement}

For the review stage, we provide a ZIP file containing the code and instructions for reviewers to reproduce and verify our results. We will release the code in a public repository and publish the trained model weights upon publication.

    \bibliography{iclr2027_conference}

@article{hudson2024comparison,
	title        = {A comparison of clustering models for inference of t cell receptor antigen specificity},
	author       = {Hudson, Dan and Lubbock, Alex and Basham, Mark and Koohy, Hashem},
	year         = 2024,
	journal      = {ImmunoInformatics},
	publisher    = {Elsevier},
	volume       = 13,
	pages        = 100033
}

@article{davis1988t,
	title        = {T-cell antigen receptor genes and T-cell recognition},
	author       = {Davis, Mark M and Bjorkman, Pamela J},
	year         = 1988,
	journal      = {Nature},
	publisher    = {Nature Publishing Group UK London},
	volume       = 334,
	number       = 6181,
	pages        = {395--402}
}

@article{bosselut2019t,
	title        = {T cell antigen recognition: Evolution-driven affinities},
	author       = {Bosselut, R{\'e}my},
	year         = 2019,
	journal      = {Proceedings of the National Academy of Sciences},
	publisher    = {National Acad Sciences},
	volume       = 116,
	number       = 44,
	pages        = {21969--21971}
}

@article{joglekar2021t,
	title        = {T cell antigen discovery},
	author       = {Joglekar, Alok V and Li, Guideng},
	year         = 2021,
	journal      = {Nature methods},
	publisher    = {Nature Publishing Group US New York},
	volume       = 18,
	number       = 8,
	pages        = {873--880}
}

@article{rojas2023personalized,
	title        = {Personalized RNA neoantigen vaccines stimulate T cells in pancreatic cancer},
	author       = {Rojas, Luis A and Sethna, Zachary and Soares, Kevin C and Olcese, Cristina and Pang, Nan and Patterson, Erin and Lihm, Jayon and Ceglia, Nicholas and Guasp, Pablo and Chu, Alexander and others},
	year         = 2023,
	journal      = {Nature},
	publisher    = {Nature Publishing Group UK London},
	volume       = 618,
	number       = 7963,
	pages        = {144--150}
}

@article{poorebrahim2021tcr,
	title        = {TCR-like CARs and TCR-CARs targeting neoepitopes: an emerging potential},
	author       = {Poorebrahim, Mansour and Mohammadkhani, Niloufar and Mahmoudi, Reza and Gholizadeh, Monireh and Fakhr, Elham and Cid-Arregui, Angel},
	year         = 2021,
	journal      = {Cancer gene therapy},
	publisher    = {Nature Publishing Group US New York},
	volume       = 28,
	number       = 6,
	pages        = {581--589}
}

@article{neefjes2011towards,
	title        = {Towards a systems understanding of MHC class I and MHC class II antigen presentation},
	author       = {Neefjes, Jacques and Jongsma, Marlieke LM and Paul, Petra and Bakke, Oddmund},
	year         = 2011,
	journal      = {Nature reviews immunology},
	publisher    = {Nature Publishing Group UK London},
	volume       = 11,
	number       = 12,
	pages        = {823--836}
}

@article{hudson2023can,
	title        = {Can we predict T cell specificity with digital biology and machine learning?},
	author       = {Hudson, Dan and Fernandes, Ricardo A and Basham, Mark and Ogg, Graham and Koohy, Hashem},
	year         = 2023,
	journal      = {Nature Reviews Immunology},
	publisher    = {Nature Publishing Group UK London},
	volume       = 23,
	number       = 8,
	pages        = {511--521}
}

@article{mayer2021tcr,
	title        = {TCR meta-clonotypes for biomarker discovery with tcrdist3 enabled identification of public, HLA-restricted clusters of SARS-CoV-2 TCRs},
	author       = {Mayer-Blackwell, Koshlan and Schattgen, Stefan and Cohen-Lavi, Liel and Crawford, Jeremy C and Souquette, Aisha and Gaevert, Jessica A and Hertz, Tomer and Thomas, Paul G and Bradley, Philip and Fiore-Gartland, Andrew},
	year         = 2021,
	journal      = {Elife},
	publisher    = {eLife Sciences Publications Limited},
	volume       = 10,
	pages        = {e68605}
}

@article{zhang2021giana,
	title        = {GIANA allows computationally-efficient TCR clustering and multi-disease repertoire classification by isometric transformation},
	author       = {Zhang, Hongyi and Zhan, Xiaowei and Li, Bo},
	year         = 2021,
	journal      = {Nature communications},
	publisher    = {Nature Publishing Group UK London},
	volume       = 12,
	number       = 1,
	pages        = 4699
}

@article{valkiers2021clustcr,
	title        = {ClusTCR: a python interface for rapid clustering of large sets of CDR3 sequences with unknown antigen specificity},
	author       = {Valkiers, Sebastiaan and Van Houcke, Max and Laukens, Kris and Meysman, Pieter},
	year         = 2021,
	journal      = {Bioinformatics},
	publisher    = {Oxford University Press},
	volume       = 37,
	number       = 24,
	pages        = {4865--4867}
}

@article{meynard2024tulip,
	title        = {TULIP: A transformer-based unsupervised language model for interacting peptides and T cell receptors that generalizes to unseen epitopes},
	author       = {Meynard-Piganeau, Barthelemy and Feinauer, Christoph and Weigt, Martin and Walczak, Aleksandra M and Mora, Thierry},
	year         = 2024,
	journal      = {Proceedings of the National Academy of Sciences},
	publisher    = {National Acad Sciences},
	volume       = 121,
	number       = 24,
	pages        = {e2316401121}
}

@article{abramson2024accurate,
	title        = {Accurate structure prediction of biomolecular interactions with AlphaFold 3},
	author       = {Abramson, Josh and Adler, Jonas and Dunger, Jack and Evans, Richard and Green, Tim and Pritzel, Alexander and Ronneberger, Olaf and Willmore, Lindsay and Ballard, Andrew J and Bambrick, Joshua and others},
	year         = 2024,
	journal      = {Nature},
	publisher    = {Nature Publishing Group UK London},
	pages        = {1--3}
}

@article{croce2024deep,
	title        = {Deep learning predictions of TCR-epitope interactions reveal epitope-specific chains in dual alpha T cells},
	author       = {Croce, Giancarlo and Bobisse, Sara and Moreno, Dana L{\'e}a and Schmidt, Julien and Guillame, Philippe and Harari, Alexandre and Gfeller, David},
	year         = 2024,
	journal      = {Nature Communications},
	publisher    = {Nature Publishing Group UK London},
	volume       = 15,
	number       = 1,
	pages        = 3211
}

@article{jensen2023nettcr,
	title        = {NetTCR 2.2-Improved TCR specificity predictions by combining pan-and peptide-specific training strategies, loss-scaling and integration of sequence similarity},
	author       = {Jensen, Mathias Fynbo and Nielsen, Morten},
	year         = 2023,
	journal      = {bioRxiv},
	publisher    = {Cold Spring Harbor Laboratory},
	pages        = {2023--10}
}

@article{leem2018stcrdab,
	title        = {STCRDab: the structural T-cell receptor database},
	author       = {Leem, Jinwoo and de Oliveira, Saulo H P and Krawczyk, Konrad and Deane, Charlotte M},
	year         = 2018,
	journal      = {Nucleic acids research},
	publisher    = {Oxford University Press},
	volume       = 46,
	number       = {D1},
	pages        = {D406--D412}
}

@article{lin2025tcr3d,
	title        = {TCR3d 2.0: expanding the T cell receptor structure database with new structures, tools and interactions},
	author       = {Lin, Valerie and Cheung, Melyssa and Gowthaman, Ragul and Eisenberg, Maya and Baker, Brian M and Pierce, Brian G},
	year         = 2025,
	journal      = {Nucleic Acids Research},
	publisher    = {Oxford University Press},
	volume       = 53,
	number       = {D1},
	pages        = {D604--D608}
}

@inproceedings{li2025rational,
	title        = {Rational Multi-Modal Transformers for TCR-pMHC Prediction},
	author       = {Li, Jiarui and Yin, Zixiang and Ding, Zhengming and Landry, Samuel J and Mettu, Ramgopal R},
	year         = 2025,
	booktitle    = {Proceedings of the 16th ACM International Conference on Bioinformatics, Computational Biology and Health Informatics},
	pages        = {1--10}
}

@misc{immrep25,
	title        = {IMMREP25: TCR Specificity Prediction Challenge},
	author       = {IMMREP25},
	year         = 2025,
	note         = {Kaggle},
	howpublished = {\url{https://kaggle.com/competitions/immrep25}}
}

@article{nielsen2024lessons,
	title        = {Lessons learned from the IMMREP23 TCR-epitope prediction challenge},
	author       = {Nielsen, Morten and Eugster, Anne and Jensen, Mathias Fynbo and Goel, Manisha and Tiffeau-Mayer, Andreas and Pelissier, Aurelien and Valkiers, Sebastiaan and Mart{\'\i}nez, Mar{\'\i}a Rodr{\'\i}guez and Meynard-Piganeeau, Barth{\'e}l{\'e}my and Greiff, Victor and others},
	year         = 2024,
	journal      = {ImmunoInformatics},
	publisher    = {Elsevier},
	volume       = 16,
	pages        = 100045
}

@article{yin2023tcrmodel2,
	title        = {TCRmodel2: high-resolution modeling of T cell receptor recognition using deep learning},
	author       = {Yin, Rui and Ribeiro-Filho, Helder V and Lin, Valerie and Gowthaman, Ragul and Cheung, Melyssa and Pierce, Brian G},
	year         = 2023,
	journal      = {Nucleic Acids Research},
	publisher    = {Oxford University Press},
	volume       = 51,
	number       = {W1},
	pages        = {W569--W576}
}

@article{deleuran2025nettcr,
	title        = {NetTCR-struc, a structure driven approach for prediction of TCR-pMHC interactions},
	author       = {Deleuran, Sebastian Nymann and Nielsen, Morten},
	year         = 2025,
	journal      = {Frontiers in Immunology},
	publisher    = {Frontiers},
	volume       = 16,
	pages        = 1616328
}

@article{li2025tcr,
	title        = {TCR-EML: Explainable Model Layers for TCR-pMHC Prediction},
	author       = {Li, Jiarui and Yin, Zixiang and Ding, Zhengming and Landry, Samuel J and Mettu, Ramgopal R},
	year         = 2025,
	journal      = {arXiv preprint arXiv:2510.04377}
}

@article{lu2026assessment,
  title={Assessment of computational methods in predicting TCR--epitope binding recognition},
  author={Lu, Yanping and Wang, Yuyan and Xu, Meng and Xie, Bingbing and Yang, Yumeng and Xu, Haodong and Suo, Shengbao},
  journal={Nature Methods},
  volume={23},
  number={1},
  pages={248--259},
  year={2026},
  publisher={Nature Publishing Group US New York}
}

@article{wu2025fast,
  title={Fast and accurate modeling of TCR-peptide-MHC complexes using tFold-TCR},
  author={Wu, Fandi and Zhao, Yu and Xiao, Yang and Qin, Chenchen and Wang, Fang and Wu, Zihan and Huang, Long-Kai and Liu, Xiao and Song, Jiangning and He, Bing and others},
  journal={bioRxiv},
  pages={2025--01},
  year={2025},
  publisher={Cold Spring Harbor Laboratory}
}

@article{shah2026unpaired,
  title={Unpaired TCR$\alpha$+ TCR$\beta$ sequencing is sufficient for training machine learning TCR-epitope recognition predictors},
  author={Shah, Aisha and Genolet, Raphael and Auger, Aymeric and Moreno, Dana L{\'e}a and Liu, Yan and Croce, Giancarlo and Racle, Julien and Harari, Alexandre and Gfeller, David},
  journal={bioRxiv},
  pages={2026--03},
  year={2026},
  publisher={Cold Spring Harbor Laboratory}
}

@article{chen2024applying,
  title={Applying interpretable machine learning in computational biology—pitfalls, recommendations and opportunities for new developments},
  author={Chen, Valerie and Yang, Muyu and Cui, Wenbo and Kim, Joon Sik and Talwalkar, Ameet and Ma, Jian},
  journal={Nature methods},
  volume={21},
  number={8},
  pages={1454--1461},
  year={2024},
  publisher={Nature Publishing Group US New York}
}

@article{jordan2015machine,
  title={Machine learning: Trends, perspectives, and prospects},
  author={Jordan, Michael I and Mitchell, Tom M},
  journal={Science},
  volume={349},
  number={6245},
  pages={255--260},
  year={2015},
  publisher={American Association for the Advancement of Science}
}

@article{shen2017deep,
  title={Deep learning in medical image analysis},
  author={Shen, Dinggang and Wu, Guorong and Suk, Heung-Il},
  journal={Annual review of biomedical engineering},
  volume={19},
  pages={221--248},
  year={2017},
  publisher={Annual Reviews}
}

@article{richards2019deep,
  title={A deep learning framework for neuroscience},
  author={Richards, Blake A and Lillicrap, Timothy P and Beaudoin, Philippe and Bengio, Yoshua and Bogacz, Rafal and Christensen, Amelia and Clopath, Claudia and Costa, Rui Ponte and de Berker, Archy and Ganguli, Surya and others},
  journal={Nature neuroscience},
  volume={22},
  number={11},
  pages={1761--1770},
  year={2019},
  publisher={Nature Publishing Group US New York}
}

@article{dwivedi2023explainable,
  title={Explainable AI (XAI): Core ideas, techniques, and solutions},
  author={Dwivedi, Rudresh and Dave, Devam and Naik, Het and Singhal, Smiti and Omer, Rana and Patel, Pankesh and Qian, Bin and Wen, Zhenyu and Shah, Tejal and Morgan, Graham and others},
  journal={ACM computing surveys},
  volume={55},
  number={9},
  pages={1--33},
  year={2023},
  publisher={ACM New York, NY}
}

@article{arrieta2020explainable,
  title={Explainable Artificial Intelligence (XAI): Concepts, taxonomies, opportunities and challenges toward responsible AI},
  author={Arrieta, Alejandro Barredo and D{\'\i}az-Rodr{\'\i}guez, Natalia and Del Ser, Javier and Bennetot, Adrien and Tabik, Siham and Barbado, Alberto and Garc{\'\i}a, Salvador and Gil-L{\'o}pez, Sergio and Molina, Daniel and Benjamins, Richard and others},
  journal={Information fusion},
  volume={58},
  pages={82--115},
  year={2020},
  publisher={Elsevier}
}

@inproceedings{
li2025quantifying,
title={Quantifying Cross-Attention Interaction in Transformers for Interpreting {TCR}-p{MHC} Binding},
author={Jiarui Li and Zixiang Yin and Haley Smith and Zhengming Ding and Samuel J Landry and Ramgopal R. Mettu},
booktitle={The Fourteenth International Conference on Learning Representations},
year={2026},
url={https://openreview.net/forum?id=S3kSOFhs5m}
}

@article{miller2019explanation,
  title={Explanation in artificial intelligence: Insights from the social sciences},
  author={Miller, Tim},
  journal={Artificial intelligence},
  volume={267},
  pages={1--38},
  year={2019},
  publisher={Elsevier}
}

@article{lipton2018mythos,
  title={The Mythos of Model Interpretability},
  author={LIPTON, ZACHARY C},
  journal={COMMUNICATIONS OF THE ACM},
  volume={61},
  number={10},
  year={2018}
}

@article{wang2023scientific,
  title={Scientific discovery in the age of artificial intelligence},
  author={Wang, Hanchen and Fu, Tianfan and Du, Yuanqi and Gao, Wenhao and Huang, Kexin and Liu, Ziming and Chandak, Payal and Liu, Shengchao and Van Katwyk, Peter and Deac, Andreea and others},
  journal={Nature},
  volume={620},
  number={7972},
  pages={47--60},
  year={2023},
  publisher={Nature Publishing Group UK London}
}

@inproceedings{kumar2020problems,
  title={Problems with Shapley-value-based explanations as feature importance measures},
  author={Kumar, I Elizabeth and Venkatasubramanian, Suresh and Scheidegger, Carlos and Friedler, Sorelle},
  booktitle={International conference on machine learning},
  pages={5491--5500},
  year={2020},
  organization={PMLR}
}

@inproceedings{selvaraju2017grad,
  title={Grad-cam: Visual explanations from deep networks via gradient-based localization},
  author={Selvaraju, Ramprasaath R and Cogswell, Michael and Das, Abhishek and Vedantam, Ramakrishna and Parikh, Devi and Batra, Dhruv},
  booktitle={Proceedings of the IEEE international conference on computer vision},
  pages={618--626},
  year={2017}
}

@inproceedings{achtibatattnlrp,
  title={AttnLRP: Attention-Aware Layer-Wise Relevance Propagation for Transformers},
  author={Achtibat, Reduan and Hatefi, Sayed Mohammad Vakilzadeh and Dreyer, Maximilian and Jain, Aakriti and Wiegand, Thomas and Lapuschkin, Sebastian and Samek, Wojciech},
  booktitle={Forty-first International Conference on Machine Learning},
  year={2024}
}

@article{gao2025weight,
  title={Weight-sparse transformers have interpretable circuits},
  author={Gao, Leo and Rajaram, Achyuta and Coxon, Jacob and Govande, Soham V and Baker, Bowen and Mossing, Dan},
  journal={arXiv preprint arXiv:2511.13653},
  year={2025}
}

@techreport{anthropic2025biology,
    author = {Lindsey, Jack and
          Gurnee, Wes and
          Ameisen, Emmanuel and
          Chen, Brian and
          Pearce, Adam and
          Turner, Nicholas L. and
          Citro, Craig and
          Abrahams, David and
          Carter, Shan and
          Hosmer, Basil and
          Marcus, Jonathan and
          Sklar, Michael and
          Templeton, Adly and
          Bricken, Trenton and
          McDougall, Callum and
          Cunningham, Hoagy and
          Henighan, Thomas and
          Jermyn, Adam and
          Jones, Andy and
          Persic, Andrew and
          Qi, Zhenyi and
          Thompson, T. Ben and
          Zimmerman, Sam and
          Rivoire, Kelley and
          Conerly, Thomas and
          Olah, Chris and
          Batson, Joshua},
  title        = {On the Biology of a Large Language Model},
  institution  = {Anthropic},
  year         = {2025},
  note         = {\url{https://transformer-circuits.pub/2025/attribution-graphs/biology.html} (Accessed: 2026-03-02)}
}

@article{jiang2021layercam,
  title={Layercam: Exploring hierarchical class activation maps for localization},
  author={Jiang, Peng-Tao and Zhang, Chang-Bin and Hou, Qibin and Cheng, Ming-Ming and Wei, Yunchao},
  journal={IEEE transactions on image processing},
  volume={30},
  pages={5875--5888},
  year={2021},
  publisher={IEEE}
}

@article{zhao2024explainability,
  title={Explainability for large language models: A survey},
  author={Zhao, Haiyan and Chen, Hanjie and Yang, Fan and Liu, Ninghao and Deng, Huiqi and Cai, Hengyi and Wang, Shuaiqiang and Yin, Dawei and Du, Mengnan},
  journal={ACM Transactions on Intelligent Systems and Technology},
  volume={15},
  number={2},
  pages={1--38},
  year={2024},
  publisher={ACM New York, NY}
}

@article{hunklinger2026towards,
  title={Towards the explainability of protein language models},
  author={Hunklinger, Andrea and Ferruz, Noelia},
  journal={Nature Machine Intelligence},
  pages={1--14},
  year={2026},
  publisher={Nature Publishing Group UK London}
}

@article{richardson2026immrep25,
  title={IMMREP25: Unseen Peptides},
  author={Richardson, Eve and Aarts, Yannick Jurriaan Maria and Altin, John A and Baakman, Coos AB and Bradley, Philip and Chen, Binbin and Clifford, Joakim and Dhar, Manjima and Diepenbroek, Danielle and Fast, Ethan and others},
  journal={bioRxiv},
  pages={2026--03},
  year={2026},
  publisher={Cold Spring Harbor Laboratory}
}

@article{li2026structure,
  title={Structure-Regularized Interpretable TCR-Epitope Prediction},
  author={Li, Jiarui and Yin, Zixiang and Zhang, Yunbei and Wang, Janet and Landry, Samuel J and Ding, Zhengming and Mettu, Ramgopal R},
  journal={arXiv preprint arXiv:2606.30902},
  year={2026}
}

@article{passaro2025boltz,
  title={Boltz-2: Towards accurate and efficient binding affinity prediction},
  author={Passaro, Saro and Corso, Gabriele and Wohlwend, Jeremy and Reveiz, Mateo and Thaler, Stephan and Somnath, Vignesh Ram and Getz, Noah and Portnoi, Tally and Roy, Julien and Stark, Hannes and others},
  journal={BioRxiv},
  year={2025}
}

@misc{openfold3-preview,
  title = {OpenFold3-preview},
  author = {{The OpenFold3 Team}},
  year = {2025},
  version = {0.4.2},
  doi = {10.5281/zenodo.19001000},
  url = {https://github.com/aqlaboratory/openfold-3}
}

@article{sanou2026imgt,
  title={IMGT{\textregistered} at scale: FAIR, dynamic, and automated tools for immune locus analysis},
  author={Sanou, Gaoussou and Zeitoun, Guilhem and Manso, Taciana and Eidi, Milad and Batool, Shamsa and Kushwaha, Anjana and Grand, Fran{\c{c}}ois and Croze, Myriam and Vaillant, Axel and Debbagh, Chahrazed and others},
  journal={Nucleic Acids Research},
  volume={54},
  number={D1},
  pages={D1119--D1132},
  year={2026},
  publisher={Oxford University Press}
}

@article{parizi2026swifttcr,
  title={SwiftTCR: Efficient computational docking protocol of TCRpMHC-I complexes using restricted rotation matrices},
  author={Parizi, Farzaneh M and Aarts, Yannick JM and Smit, Nils and Dona Roran, AR and Diepenbroek, Dani{\"e}lle and Kr{\"o}sschell, Wieke A and Thijs, Levin and Tepperik, Joost and Marzella, Dario F and Ramakrishnan, Gayatri and others},
  journal={Briefings in Bioinformatics},
  volume={27},
  number={3},
  pages={bbag306},
  year={2026},
  publisher={Oxford University Press}
}
    \bibliographystyle{styles/iclr2027_conference}
    \newpage
\appendix
\section{Appendix}

\subsection{TCR-Epitope Binding Introduction}
TCR-epitope recognition is a key step in T cell activation and adaptive immune responses. Antigens are first taken up by antigen-presenting cells (APCs) and processed into peptides. These peptides bind to MHC molecules to form peptide-MHC (pMHC) complexes, which are presented on the APC surface and recognized by $\alpha\beta$-TCRs on T cells~\citep{davis1988t,neefjes2011towards}. MHC molecules are generally classified into MHC-I and MHC-II, which primarily present intracellular antigens recognized by CD4+ T cell, such as tumor-associated antigens, and exogenous antigens recognized by CD8 T cell, such as viral antigens, respectively~\citep{hudson2023can}. An $\alpha\beta$-TCR consists of one alpha chain and one beta chain, each containing three complementarity-determining regions (CDRs). These regions, particularly CDR3, play important roles in determining whether a TCR recognizes a pMHC complex. The six CDRs are CDR1a, CDR2a, CDR3a, CDR1b, CDR2b, and CDR3b~\citep{bosselut2019t}.

\subsection{TCR-Epitope Prediction Models.}
TCR-epitope prediction models can be broadly categorized into unsupervised and supervised approaches~\citep{hudson2024comparison}. Unsupervised approaches primarily rely on sequence similarity and clustering, with representative methods including TCRdist3~\citep{mayer2021tcr}, GIANA~\citep{zhang2021giana}, and ClusTCR~\citep{valkiers2021clustcr}. Recent advances have increasingly shifted toward supervised learning, with models adopting diverse neural architectures. These approaches include CNNs, such as NetTCR-2.2~\citep{jensen2023nettcr}, and transformer-based models, such as TULIP~\citep{meynard2024tulip}, MixTCRpred~\citep{croce2024deep}, and TCR-SRIM~\citep{li2026structure}. NetTCR-2.2 and MixTCRpred concatenate CDR regions with the peptide sequence for binding prediction. TULIP uses CDR3 regions and peptide sequences, with an optional MHC allele as input. TCR-SRIM uses CDR3 regions and peptide sequences to predict TCR-epitope binding.

\subsection{Model Interpretability.}
Deep learning models can be broadly categorized into black-box and interpret-by-design models. Interpret-by-design models, such as TCR-SRIM, incorporate interpretable components directly into their architectures. In contrast, black-box models can be analyzed using post-hoc interpretability methods, which estimate the contribution of input features to model predictions. These methods can be broadly categorized into model-agnostic and model-specific approaches. SHAP is a representative model-agnostic method that estimates feature contributions based on Shapley values~\citep{kumar2020problems}. Model-specific methods are designed to exploit the architectures or internal representations of particular model families. For CNNs, class activation map (CAM) methods, including Grad-CAM~\citep{selvaraju2017grad} and LayerCAM~\citep{jiang2021layercam}, are commonly used to identify spatially important features. For transformer-based models, layer-wise relevance propagation (LRP), such as AttnLRP~\citep{achtibatattnlrp}, and attention-based methods, such as QCAI~\citep{li2025quantifying}, have been used to analyze the contribution of individual tokens or interactions to model predictions.

\subsection{TCR-XAI2 Benchmark}
\begin{figure*}[th]
    \centering
    \includegraphics[width=\linewidth]{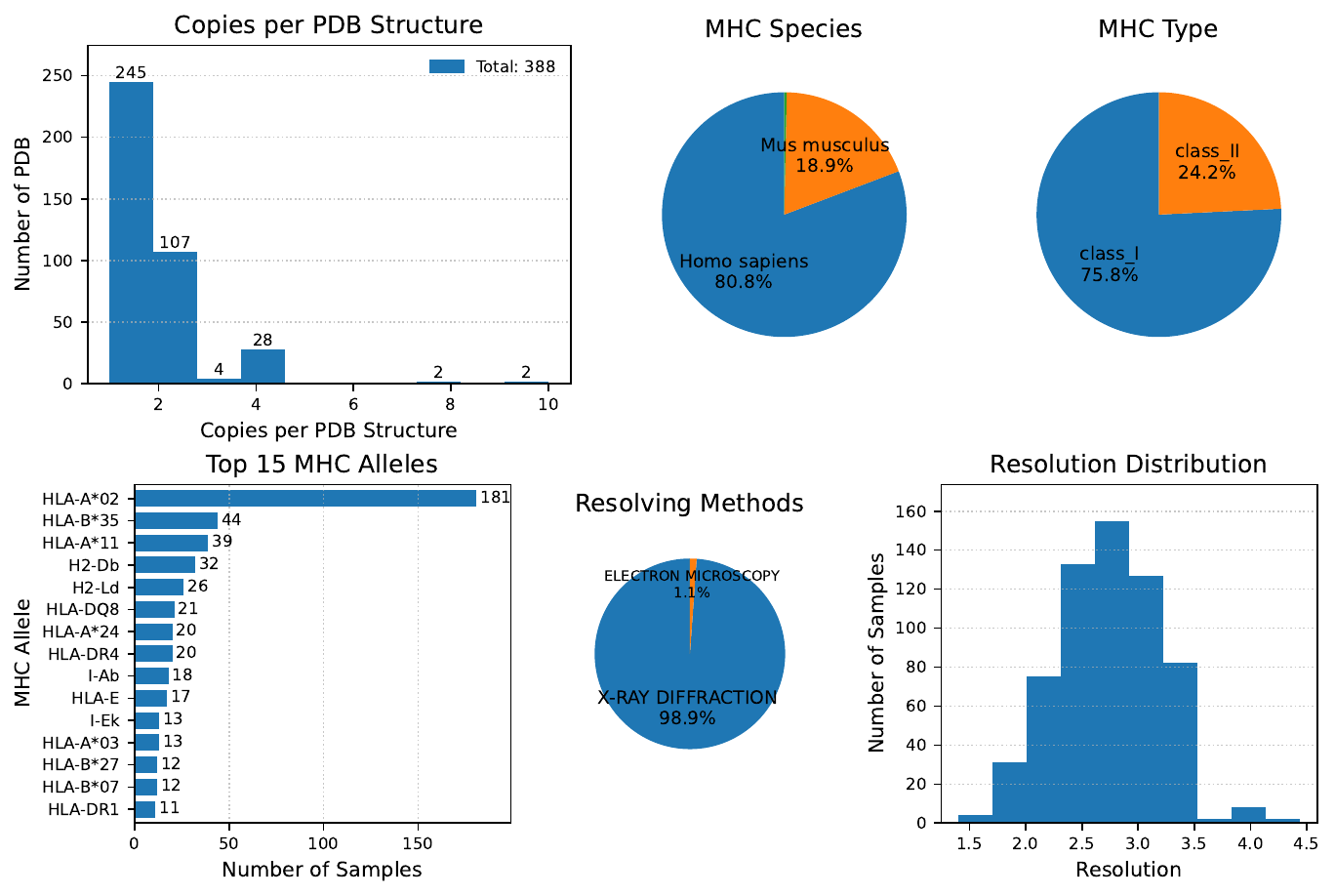}
    \caption{\textbf{The statistics of sample distribution of TCR-XAI2 benchmark.} The TCR-XAI2 benchmark consists of 388 unique structures and 619 samples, while a structure can provide multiple TCR-epitope copies. Most of the samples are from human (80.8\%), MHC-I (75.8\%), and resolved by X-ray diffraction (98.9\%).}
    \label{fig:TCR-XAI:stats}
\end{figure*}

\cite{li2025quantifying} introduced the TCR-XAI benchmark, which was compiled from experimentally resolved crystal structures to quantitatively evaluate interpretation quality. It contains 274 samples and uses residue-level contact maps as ground-truth explanations. Since TCR3D2.0~\citep{lin2025tcr3d} is continuously updated with newly resolved TCR-epitope structures, we collected the latest structures from TCR3D2.0 together with structures from STCRDab~\citep{leem2018stcrdab} to construct the TCR-XAI2 benchmark. This process yielded 388 unique structures. Because a single PDB entry may contain multiple copies of the same TCR-epitope complex, we treated each copy as an independent sample to increase the number of samples of the benchmark, resulting in 619 samples in total. Duplicate copies are shown in Fig.~\ref{fig:TCR-XAI:stats}.

Among the TCR-XAI2 samples, 80.8\% contain human MHC molecules, 18.9\% contain mouse MHC molecules, and 0.3\% contain MHC molecules from other species. Of these samples, 75.8\% involve MHC-I and 24.2\% involve MHC-II. The distribution of MHC alleles is highly imbalanced. The three most frequent human alleles are \texttt{HLA-A*02} (181 samples), \texttt{HLA-B*35} (44 samples), and \texttt{HLA-A*11} (39 samples), consistent with the allele distribution commonly observed in sequence-based TCR-epitope datasets. The two most frequent mouse alleles are \texttt{H2-Db} (32 samples) and \texttt{H2-Ld} (26 samples). Most samples (98.9\%) were resolved using X-ray diffraction, while the remaining samples were resolved using cryo-electron microscopy (cryo-EM). The structures have a mean resolution in the range of 2.5 to 3.0~\AA.

\begin{figure}
    \centering
    \includegraphics[width=\linewidth]{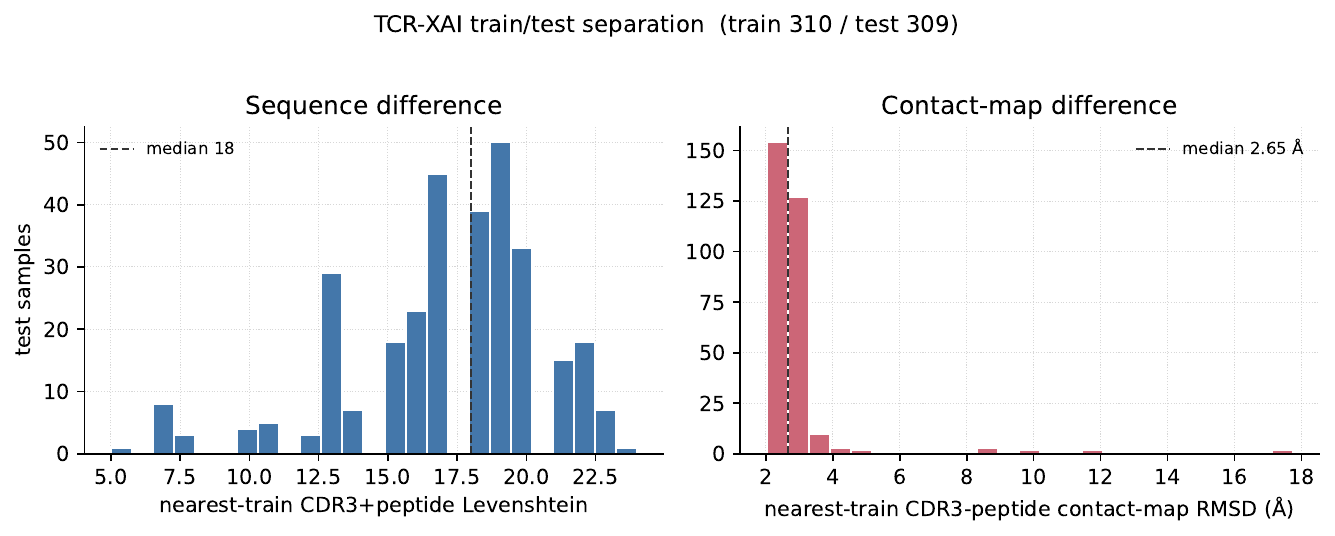}
    \caption{\textbf{The sequence and structural differences between the training and test splits of the TCR-XAI2 benchmark.} From the sequence perspective, the median distance between the training and test splits is 18 residues, with a minimum distance of 5 residues. From the structural perspective, the CDR3-peptide contact maps exhibit a median difference of at least 2.65~\AA between the training and test splits.}
    \label{fig:TCR-XAI:tts:diff}
\end{figure}

We also provide training and test splits of TCR-XAI2 for models such as TCR-SRIM, which include variants that require structural information during training. To minimize potential data leakage, we designed the split to account for differences in both sequence and structure, specifically the CDR3b-peptide and CDR3a-peptide contact maps. As shown in Fig.~\ref{fig:TCR-XAI:tts:diff}, we evaluated the minimum Levenshtein distance of CDR3 regions and peptides between the training and test splits. The median distance is 18 residues, with a minimum of 5 residues, indicating substantial sequence differences between the two splits. We further evaluated structural differences using the CDR3a-peptide and CDR3b-peptide contact maps. The median average contact distance is 2.65~\AA. These results indicate that the training and test splits are separated in both sequence and structural space, reducing the potential for data leakage from either source.

\paragraph{Predicted Structures.}
To further investigate how predicted structures influence model explanations and training, we used AlphaFold3~\citep{abramson2024accurate}, Boltz-2~\citep{passaro2025boltz}, TCRModel2~\citep{yin2023tcrmodel2}, OpenFold3~\citep{openfold3-preview}, and tFold-TCR~\citep{wu2025fast} to predict the structures corresponding to the TCR-XAI2 benchmark. For AlphaFold3 and TCRModel2, we used their official web servers to generate predictions. We deployed tFold-TCR locally, while Boltz-2 and OpenFold3 predictions were generated through the NVIDIA NIM API. As shown in Table~\ref{tab:TCR-XAI:pred:stats}, we divided each predicted-structure dataset into training and test splits using the same splitting criteria as the experimentally resolved structures. Except for AlphaFold3 and TCRModel2, all methods successfully generated structures for the full set of 619 samples. AlphaFold3 and TCRModel2 failed to generate predictions for 15 and 28 samples, respectively. In addition, TCRModel2 models only the TCR variable domains and the peptide-binding groove, omitting $\beta_2$-microglobulin, the TCR constant domains, and the MHC $\alpha_3$ domain.

\begin{table}[th]
\centering
\caption{\textbf{The number of samples and training and test splits in the predicted TCR-XAI2 benchmarks.} AlphaFold3 and TCRModel2 failed to predict 15 and 28 samples, respectively, while all other methods successfully predicted structures for all samples.}
\label{tab:TCR-XAI:pred:stats}
\begin{tabular}{lrrrr}
\toprule
\textbf{Model} & \textbf{Train} & \textbf{Test} & \textbf{Total} & \textbf{Missing} \\
\midrule
TCR-XAI2 (AlphaFold3)              & 303 & 301 & 604 & 15 \\
TCR-XAI2 (Boltz-2)                 & 310 & 309 & 619 & 0 \\
TCR-XAI2 (OpenFold)                & 310 & 309 & 619 & 0 \\
TCR-XAI2 (TCRModel2)               & 298 & 293 & 591 & 28 \\
TCR-XAI2 (tFold-TCR)               & 310 & 309 & 619 & 0 \\
\midrule
TCR-XAI2                    & 310 & 309 & 619 & - \\
\bottomrule
\end{tabular}
\end{table}

We further evaluated the quality of the predicted structures using full-structure RMSD and contact-map RMSD for the CDR3a-peptide and CDR3b-peptide interfaces, with the experimentally resolved structures as references. As shown in Table~\ref{tab:TCR-XAI:pred:stats}, TCRModel2 is not directly comparable in terms of full-structure RMSD because it does not provide the complete complex structure. Boltz-2 achieves the lowest RMSD among the evaluated methods, with CDR3-peptide interface RMSDs below 2.5~\AA and full-structure RMSDs below 3.5~\AA. AlphaFold3 and TCRModel2 also achieve CDR3-peptide contact-map RMSDs below 3.5~\AA, suggesting that their predicted interfaces remain reasonably close to the experimentally resolved structures. In contrast, tFold-TCR and OpenFold3 show substantially larger CDR3-peptide contact-map RMSDs, exceeding 6~\AA and 15~\AA, respectively.

\begin{table}[th]
\centering
\caption{\textbf{The structural quality of predictions from different models.} We compare the mean RMSD and standard deviation of the full structure and CDR3-peptide interfaces between predicted and experimentally resolved structures. Full-structure RMSD is not available for TCRModel2 because it omits several components of the complete TCR-pMHC complex.}
\label{tab:TCR-XAI:pred:quality}
\begin{tabular}{lccc}
\toprule
Predictor & CDR3a-peptide (\AA) & CDR3b-peptide (\AA) & Full (\AA) \\
\midrule
Boltz-2     & $\mathbf{2.48 \pm 7.78}$ & $\mathbf{2.38 \pm 7.59}$ & $\mathbf{3.24 \pm 5.88}$ \\
AlphaFold3  & $2.97 \pm 7.88$          & $2.95 \pm 7.74$          & $3.75 \pm 6.01$ \\
TCRmodel2   & $3.38 \pm 7.98$          & $3.35 \pm 7.74$          & - \\
tFold-TCR   & $6.35 \pm 7.54$          & $7.28 \pm 6.89$          & $10.29 \pm 7.48$ \\
OpenFold3   & $23.47 \pm 13.56$        & $16.47 \pm 8.14$         & $35.44 \pm 3.30$ \\
\bottomrule
\end{tabular}
\end{table}

\paragraph{Proxy Resolution for Predicted Structures.}
To estimate a proxy resolution for predicted structures, we compute a score from the predicted confidence of the CDR3a, CDR3b, and peptide regions following~\citep{li2026structure}. AlphaFold3, Boltz-2, OpenFold3, and TCRModel2 provide per-residue pLDDT scores $\phi_i\in[0,100]$. For each region $c\in{\text{CDR3}a,\text{CDR3}b,\text{Peptide}}$, we first compute its mean pLDDT:
\begin{equation}
\bar{\phi}_{c} = \frac{1}{|c|} \sum_{i\in c}\phi_i.
\end{equation}
We then use the minimum regional confidence as the overall confidence of the complex:
\begin{equation}
\phi_{\min} =
\min\left(
\bar{\phi}*{\mathrm{CDR3}\alpha},
\bar{\phi}*{\mathrm{CDR3}\beta},
\bar{\phi}*{\mathrm{Peptide}}
\right).
\end{equation}
Finally, we convert this confidence score to a proxy resolution:
\begin{equation}
r=\frac{100-\phi*{\min}}{10}.
\end{equation}
For tFold-TCR, whose pLDDT scores are normalized to $[0,1]$, we instead compute
\begin{equation}
r=10(1-\phi_{\min}).
\end{equation}
Thus, lower $r$ indicates higher predicted structural confidence.

\paragraph{Binding Region Hit Rate.}
To quantitatively assess the quality of model interpretations and attention-based importance scores, we adopt the binding region hit rate (BRHR) introduced by \cite{li2025quantifying}. BRHR measures the agreement between residue importance scores and the structural proximity of residues between two interacting chains, such as the peptide and CDR3b. Given a percentile threshold $t \in (0,1]$, we select the top $t$ fraction of residues with the highest importance scores in $\mathbf{S}$. A selected residue is considered a \emph{hit} if its interaction distance also falls within the top $t$ fraction of structurally interacting residues. We calculate the hit rate separately for each input sequence type in each sample and average the results across the TCR-XAI2 benchmark to obtain the final BRHR. A higher BRHR indicates better agreement between the model interpretation and the structurally defined binding regions. Following \cite{li2025quantifying}, we report BRHR@.25, which measures the hit rate among the top 25\% of residues with the highest importance scores and the 25\% nearest residues based on structural proximity.

\subsection{TCR-Epitope Prediction Models}
To investigate how different input modalities and model designs affect TCR-epitope model learning, we analyze four models using \ModelName[abbr] including NetTCR-2.2~\citep{jensen2023nettcr}, MixTCRpred~\citep{croce2024deep}, TULIP~\citep{meynard2024tulip}, and TCR-SRIM~\citep{li2026structure}. NetTCR-2.2 is a CNN-based black-box model, for which we use Grad-CAM~\citep{selvaraju2017grad} for post-hoc interpretation. MixTCRpred is an encoder-only Transformer model, and we use AttnLRP~\citep{achtibatattnlrp} to interpret its attention-based evidence. Since TULIP is an encoder-decoder Transformer, we use QCAI~\citep{li2025quantifying} to interpret its cross-attention. TCR-SRIM is an interpret-by-design model that directly provides contact maps between the peptide and CDR3a and between the peptide and CDR3b.

To examine whether the choice of post-hoc interpretation method affects the resulting explanations, we additionally apply AttnLRP to TULIP and Grad-CAM to MixTCRpred. A detailed analysis of how different post-hoc methods affect model interpretations is provided in the appendix. For the main analysis, we use the interpretation method that best matches the architecture of each model.
For MixTCRpred, we use the model variant that includes the peptide as part of the input. TULIP provides two variants, with MHC allele information (TULIP-MHC) and without MHC allele information (TULIP-NoMHC). We analyze both variants to investigate how MHC allele information affects model explanations. TCR-SRIM is evaluated without structure regularization to provide a consistent comparison with the other sequence-based models, using ESM2-8M as the backbone for computational efficiency. We additionally analyze structure-regularized TCR-SRIM (ESM2-8M) using different predicted structures to investigate how structural information affects model learning.

\begin{table}[th]
\centering
\caption{\textbf{The final performance comparison of all models on IMMREP23, IMMREP25, and their respective original test datasets.} ROC-AUC at a false positive rate below 0.1 (ROC-AUC@FPR$<0.1$) indicates that all models converged to performance comparable to that reported in their original publications during our retraining.
}
\label{tab:model:finalperf}
\begin{tabular}{lcccc}
\toprule
\multirow{2}{*}{Model} & \multicolumn{3}{c}{ROC-AUC@FPR$<0.1$} & ROC-AUC \\
\cmidrule(lr){2-4}
 & IMMREP23 & IMMREP25 & Original Test & Original Test \\
\midrule
MixTCRpred  & 0.611 & 0.513 & 0.746 & 0.850 \\
NetTCR-2.2  & 0.575 & 0.516 & 0.748 & 0.886 \\
TULIP-MHC   & 0.574 & 0.745 & 0.607 & 0.710 \\
TULIP-NoMHC & 0.587 & 0.744 & 0.632 & 0.736 \\
TCR-SRIM    & 0.580 & 0.592 & 0.624 & 0.641 \\
\bottomrule
\end{tabular}
\end{table}

For training, we train each model three times using different random seeds (0, 1, and 2). Each model is trained on its original training dataset following the default training configuration specified by its original implementation. MixTCRpred is trained for 50 epochs, with interpretation tracking performed every epoch. NetTCR-2.2 is trained for 200 epochs, with interpretations tracked every 4 epochs. TULIP is trained for 300 epochs, with interpretations tracked every 6 epochs for both the MHC and NoMHC variants. TCR-SRIM is trained for 200 epochs, with interpretations tracked every 4 epochs for both the structure-regularized and non-regularized variants.

We primarily discuss the training trajectory with seed 0 in the main text and analyze variations across random seeds in the appendix. The final performance on IMMREP23~\citep{nielsen2024lessons}, IMMREP25~\citep{immrep25}, and the test datasets provided by the respective models is reported in Table~\ref{tab:model:finalperf}. We report both ROC-AUC and ROC-AUC at a false positive rate below 0.1 (ROC-AUC@FPR$<0.1$), following the evaluation protocol used by IMMREP. We do not include the Assessment benchmark~\citep{lu2026assessment} because the evaluated models require different input modalities, resulting in two distinct groups of models that are not directly comparable.
The results show that all models converge to performance levels consistent with those reported in their respective original papers.

\subsection{Training Random Seed Variants}
\label{sec:diffrunseed}
\begin{figure}[t]
    \centering
    \includegraphics[width=0.49\linewidth]{figures/modelexplains/nettcr2.2_gradcam_real_struct_pan_0.pdf}
    \includegraphics[width=0.49\linewidth]{figures/modelexplains/tulip_qcai_nomhc_real_struct_pan_0.pdf}
    \includegraphics[width=0.49\linewidth]{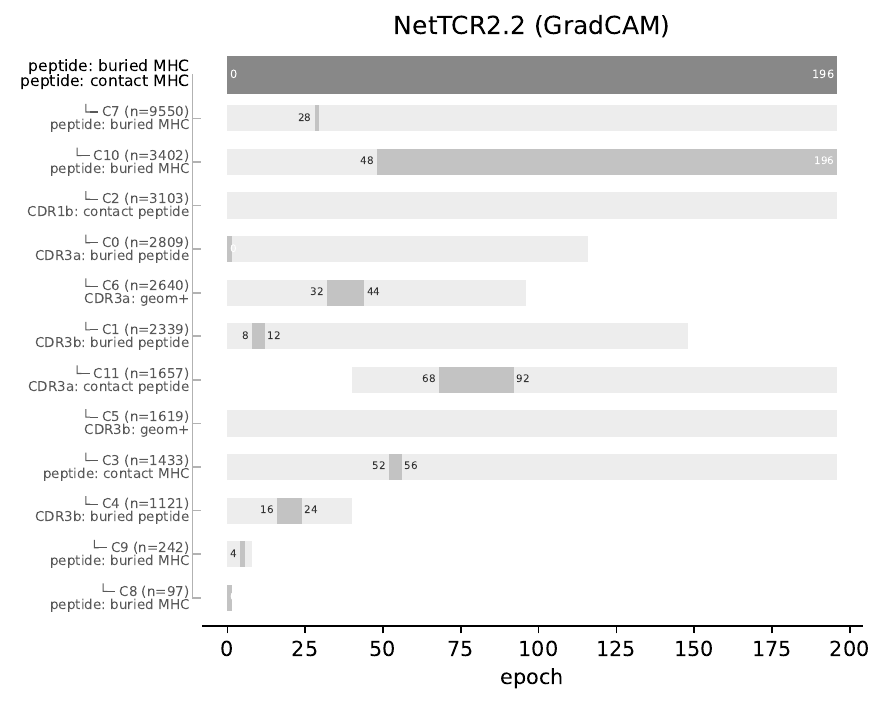}
    \includegraphics[width=0.49\linewidth]{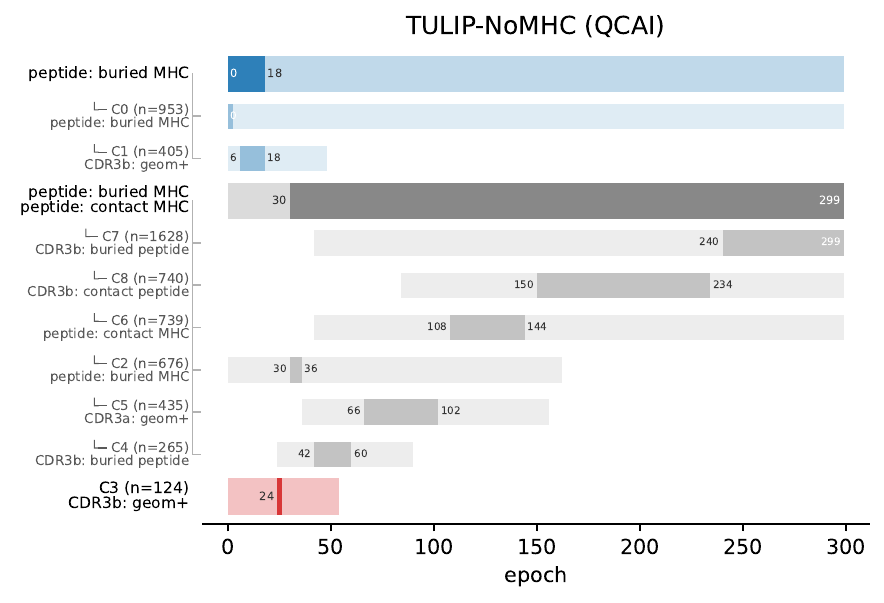}
    \includegraphics[width=0.49\linewidth]{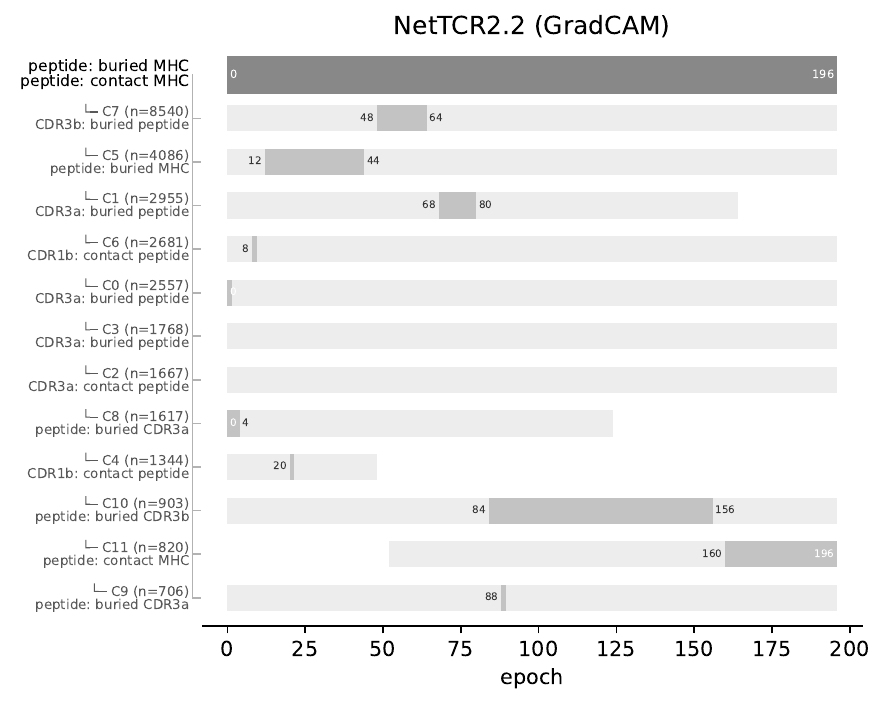}
    \includegraphics[width=0.49\linewidth]{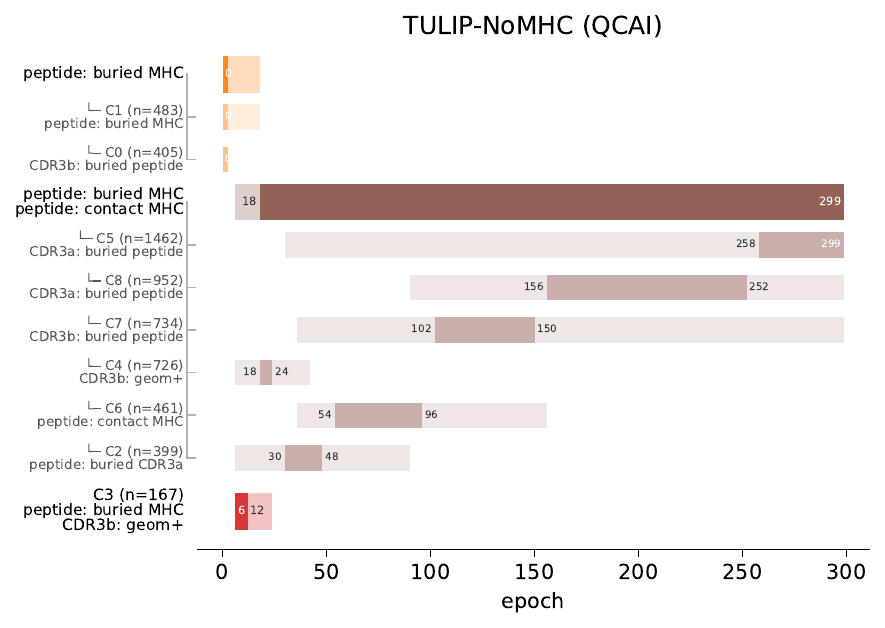}
    \caption{Different random seeds run for TULIP-NoMHC with QCAI and NetTCR-2.2 with GradCAM.}
    \label{fig:seed:variant:tulipmhc:nomhc}
\end{figure}

As model learning trajectories may vary across random seeds, we compare the explanations obtained from three random seeds for TULIP-NoMHC with QCAI and NetTCR-2.2 with GradCAM. As shown in Fig.~\ref{fig:seed:variant:tulipmhc:nomhc}, NetTCR-2.2 exhibits consistent explanations across seeds, primarily focusing on peptide-MHC interactions and, among the CDR regions, mainly CDR3a, CDR3b, and CDR1b. TULIP-NoMHC also shows similar explanations across seeds. Notably, all seeds exhibit a cluster corresponding to the CDR3b geometric region during the early stages of training. These results suggest that the models learn similar interpretation trajectories despite differences in random initialization.

\subsection{Interpretation Method Difference}
\label{sec:diffinterpmethod}
\begin{figure}[th]
    \centering
    \includegraphics[width=0.49\linewidth]{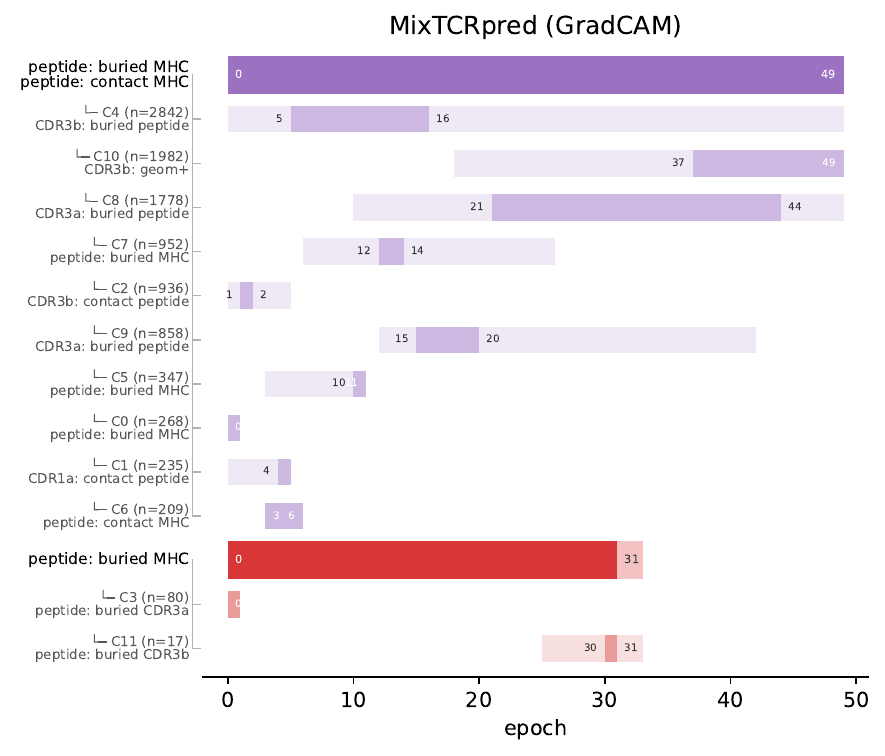}
    \includegraphics[width=0.49\linewidth]{figures/modelexplains/mixtcrpred_attnlrp_real_struct_pan_0.pdf}
    \caption{\ModelName[abbr] explanation for MixTCRpred interpreted by GradCAM and AttnLRP respectively.}
    \label{fig:seed:interp:diff:gradcamattnlrp}
\end{figure}

\begin{figure}[th]
    \centering
    \includegraphics[width=0.49\linewidth]{figures/modelexplains/tulip_qcai_mhc_real_struct_pan_0.pdf}
    \includegraphics[width=0.49\linewidth]{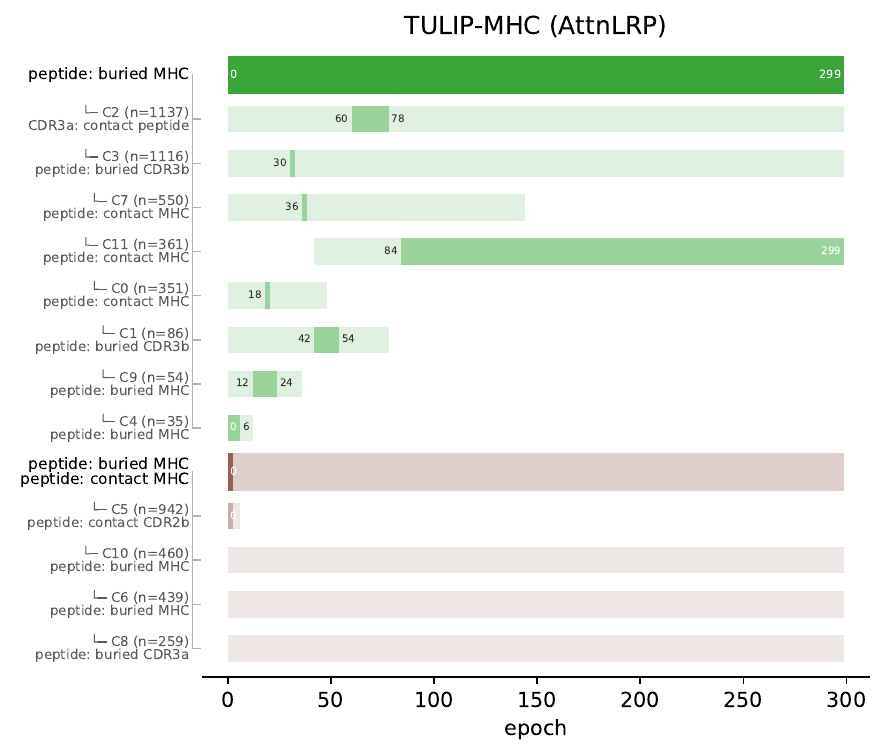}
    \caption{\ModelName[abbr] explanation for MHC-TULIP interpreted by QCAI and AttnLRP respectively.}
    \label{fig:seed:interp:diff:qcaiattnlrp}
\end{figure}

We further examine how different interpretation methods affect \ModelName[abbr] explanations for the same model. As shown in Fig.~\ref{fig:seed:interp:diff:gradcamattnlrp} and Fig.~\ref{fig:seed:interp:diff:qcaiattnlrp}, we compare GradCAM and AttnLRP for MixTCRpred, and QCAI and AttnLRP for TULIP-MHC. The results show that different interpretation methods can produce different explanations for the same model, potentially due to differences in interpretation quality. Therefore, we primarily analyze each model using its corresponding interpretation method that provides the most reliable explanations.


\subsection{Different Allele-Specific Subsets}
\begin{figure}
    \centering
    \includegraphics[width=0.55\linewidth]{figures/modelexplains/mixtcrpred_attnlrp_real_struct_HLA-A_02_0.pdf}
    \includegraphics[width=0.55\linewidth]{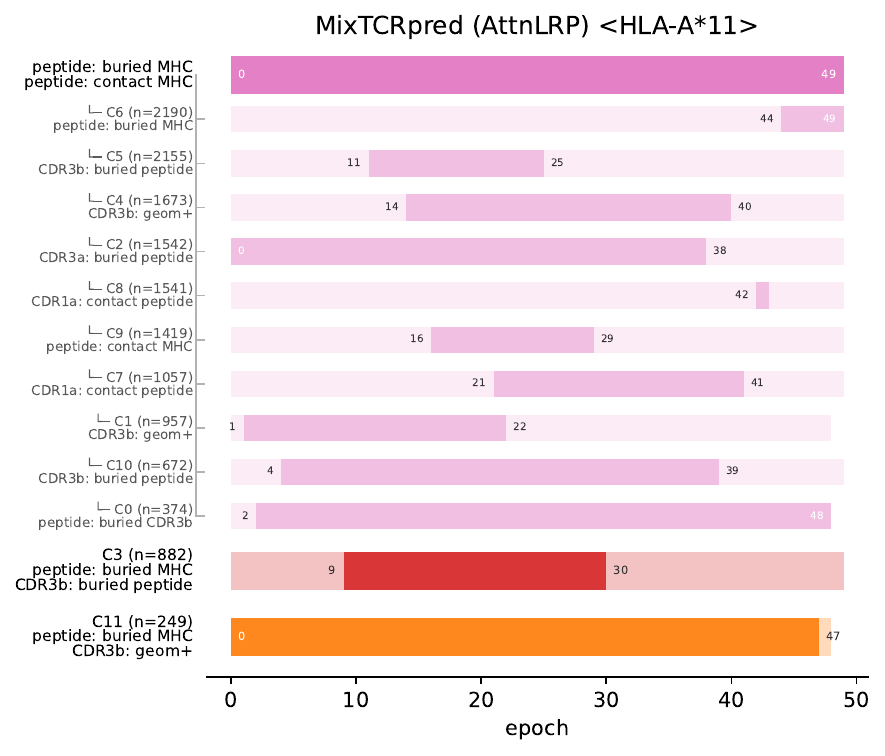}
    \includegraphics[width=0.55\linewidth]{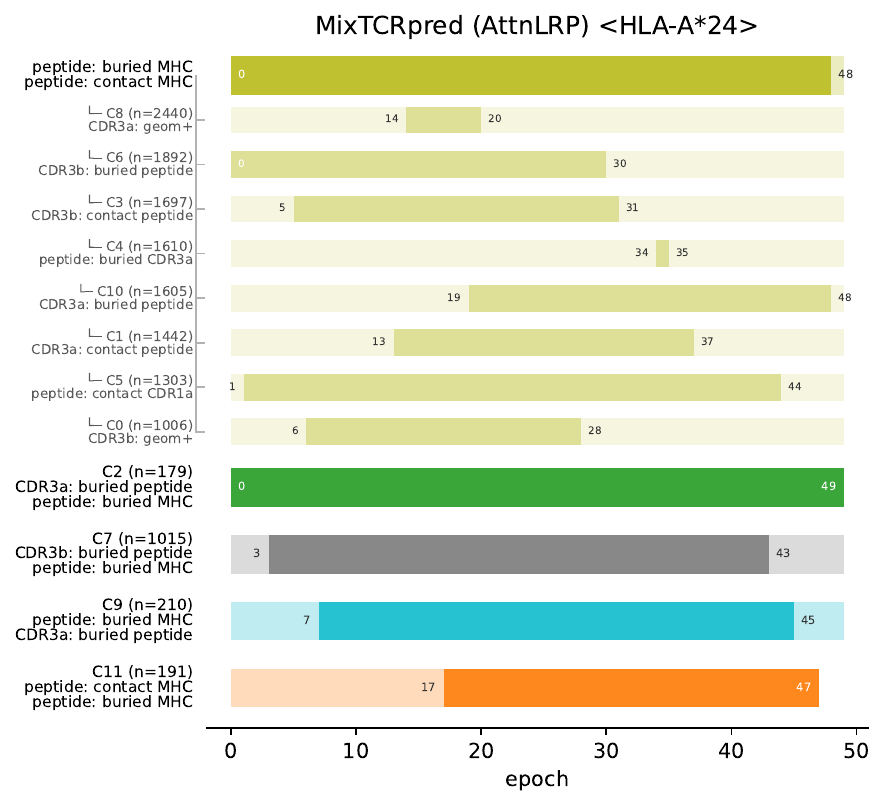}
    \caption{The \ModelName[abbr] explanation for MixTCRpred interpreted by AttnLRP on \texttt{HLA-A*02}, \texttt{HLA-A*11}, and \texttt{HLA-A*24} specific subset.}
    \label{fig:seed:mixtcrpred:diffalleles}
\end{figure}

As we observe that training TCR-epitope models on allele-specific subsets may reduce conflicts between TCR $\alpha$- and $\beta$-chain information and enable the model to learn more diverse information, we further compare models trained on different allele-specific subsets to examine this observation. As shown in Fig.~\ref{fig:seed:mixtcrpred:diffalleles}, the learning trajectories across all three alleles exhibit similar patterns. These results suggest that allele-specific training may enable models to learn from a broader range of input features and coordinate information from the TCR $\alpha$ and $\beta$ chains, thereby reducing conflicts between them.

\subsection{Explain Predicted Structure Difference in Guiding Model Training}
\begin{figure}[ht]
    \centering
    \includegraphics[width=0.5\linewidth]{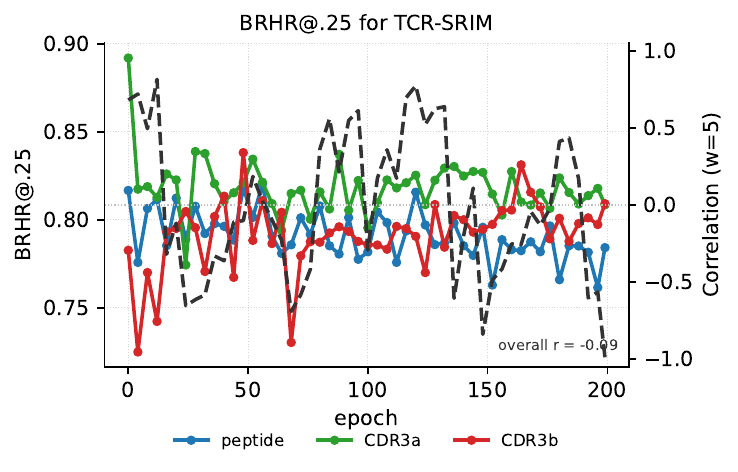}
    \caption{\textbf{The evolution of BRHR@0.25 during training for TCR-SRIM without structural regularization.} The model's interpretation remains highly unstable throughout training, with substantial fluctuations in BRHR across the three interaction regions.}
    \label{fig:tcrsrim:noreg:brhr}
\end{figure}

Recent studies have explored incorporating structural information into TCR-epitope prediction models~\citep{deleuran2025nettcr,li2026structure}. Motivated by these approaches, we use the TCR-XAI2 benchmark with structures predicted by different structure prediction models to investigate how structural quality affects model learning. As a baseline, we first examine the evolution of BRHR in TCR-SRIM without structural regularization. As shown in Fig.~\ref{fig:tcrsrim:noreg:brhr}, the BRHR trajectories of all chains fluctuate substantially throughout training rather than converging to stable patterns, indicating that the model's structural evidence remains unstable when no structural guidance is provided.

\begin{figure}[th]
    \centering
    \includegraphics[width=0.32\linewidth]{figures/brhrdeltasrim/brhrdelta_tcr-srim_real_0.pdf}
    \includegraphics[width=0.32\linewidth]{figures/brhrdeltasrim/brhrdelta_tcr-srim_alphafold3_0.pdf}
    \includegraphics[width=0.32\linewidth]{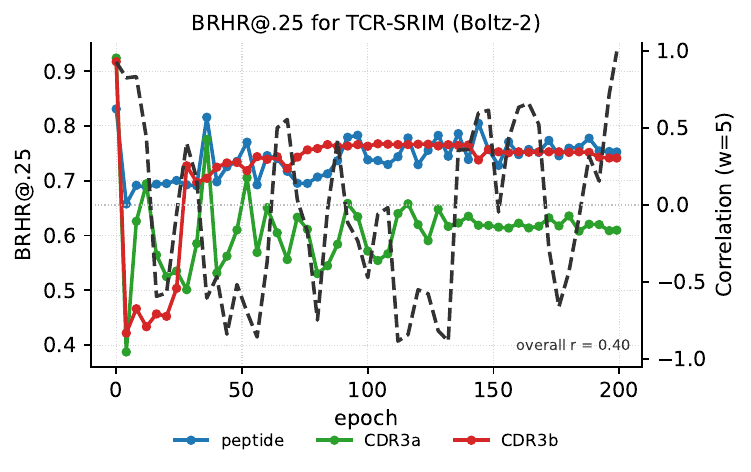}
    \includegraphics[width=0.32\linewidth]{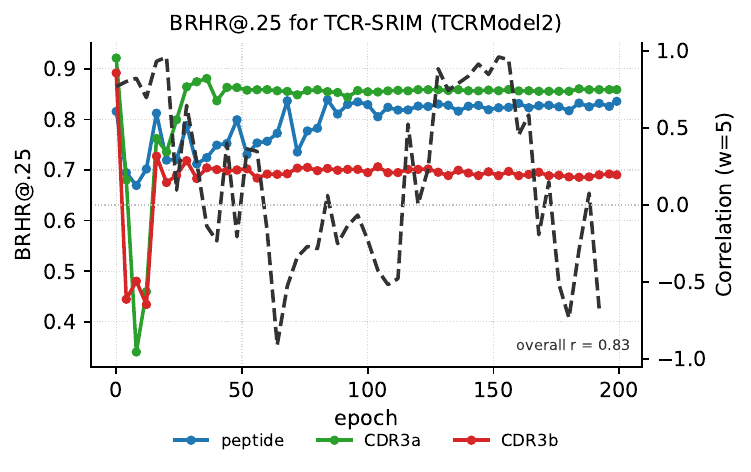}
    \includegraphics[width=0.32\linewidth]{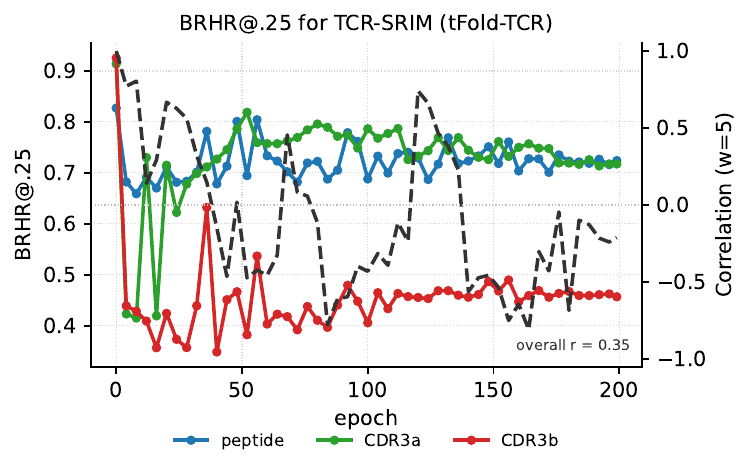}
    \includegraphics[width=0.32\linewidth]{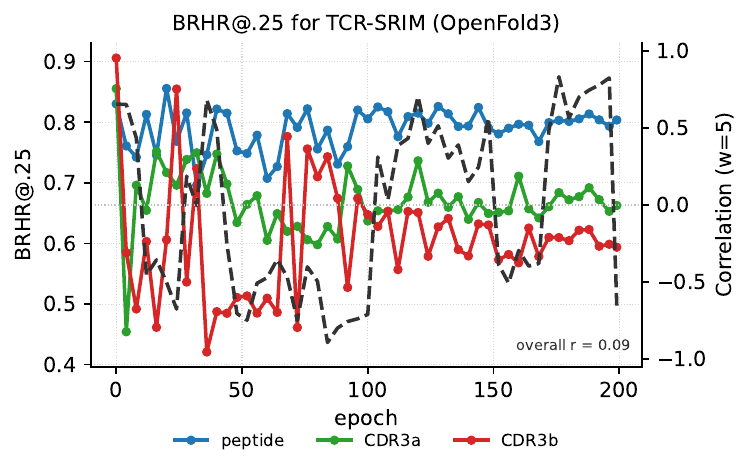}
    \caption{\textbf{The evolution of BRHR@0.25 during training for TCR-SRIM regularized with experimentally resolved structures and structures predicted by AlphaFold3, Boltz-2, TCRModel2, tFold-TCR, and OpenFold3.} Structural regularization stabilizes the model's interpretation of CDR3-peptide interactions, with the degree of stabilization varying according to the quality of the structural guidance.}
    \label{fig:srim:brhr:full}
\end{figure}

\begin{figure}[th]
    \centering
    \includegraphics[width=0.49\linewidth]{figures/modelexplains_srim/tcr-srim_experimental_pan_0.pdf}
    \includegraphics[width=0.49\linewidth]{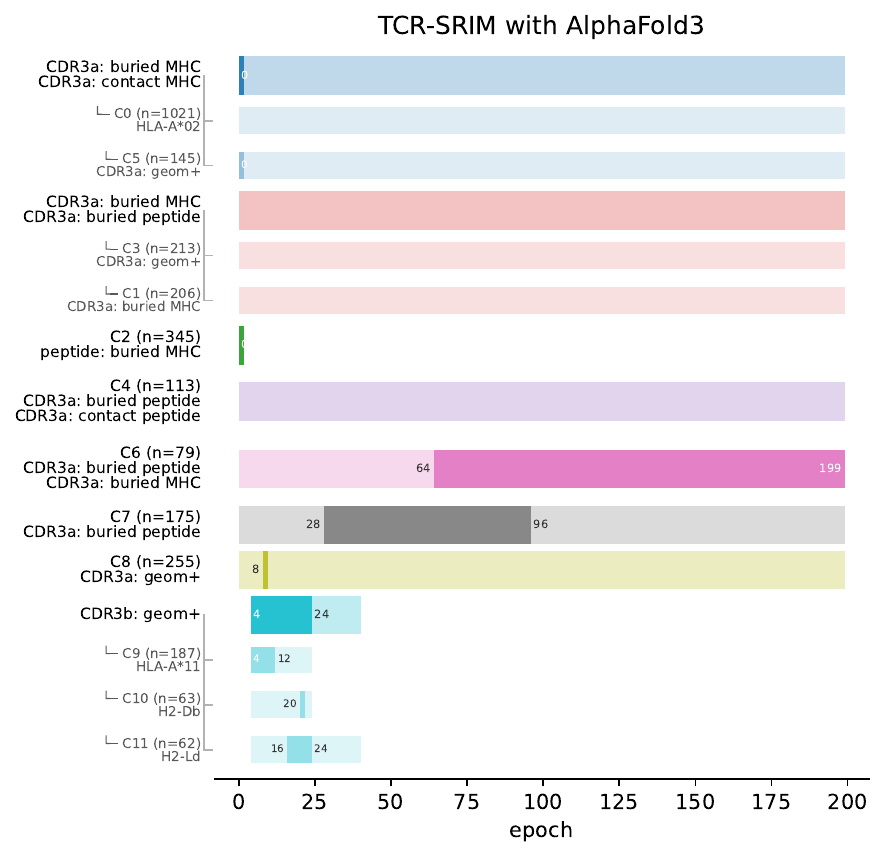}
    \includegraphics[width=0.49\linewidth]{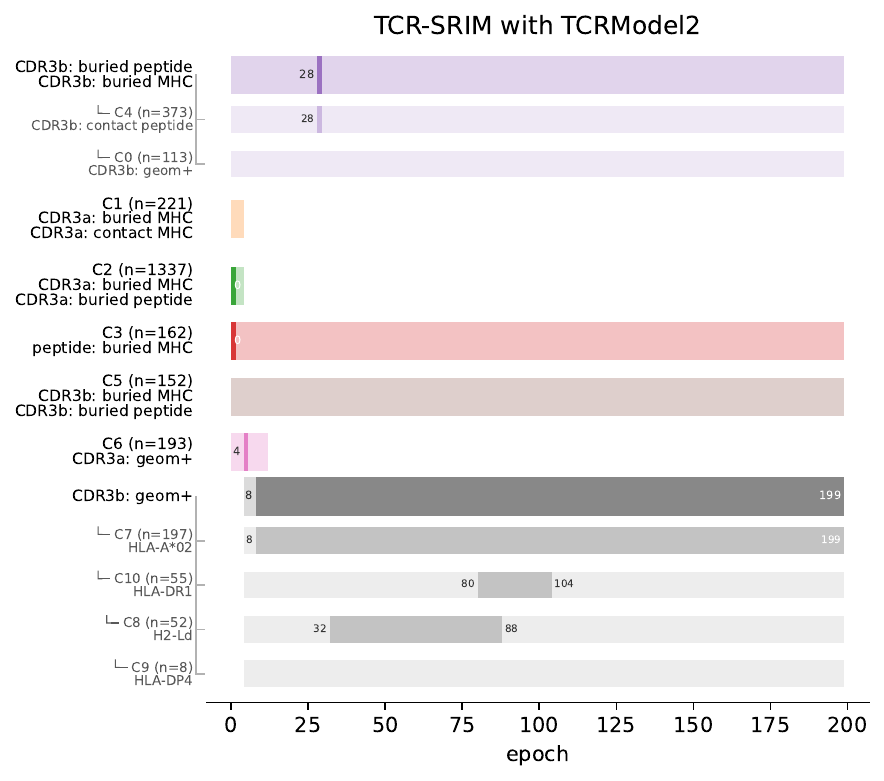}
    \includegraphics[width=0.49\linewidth]{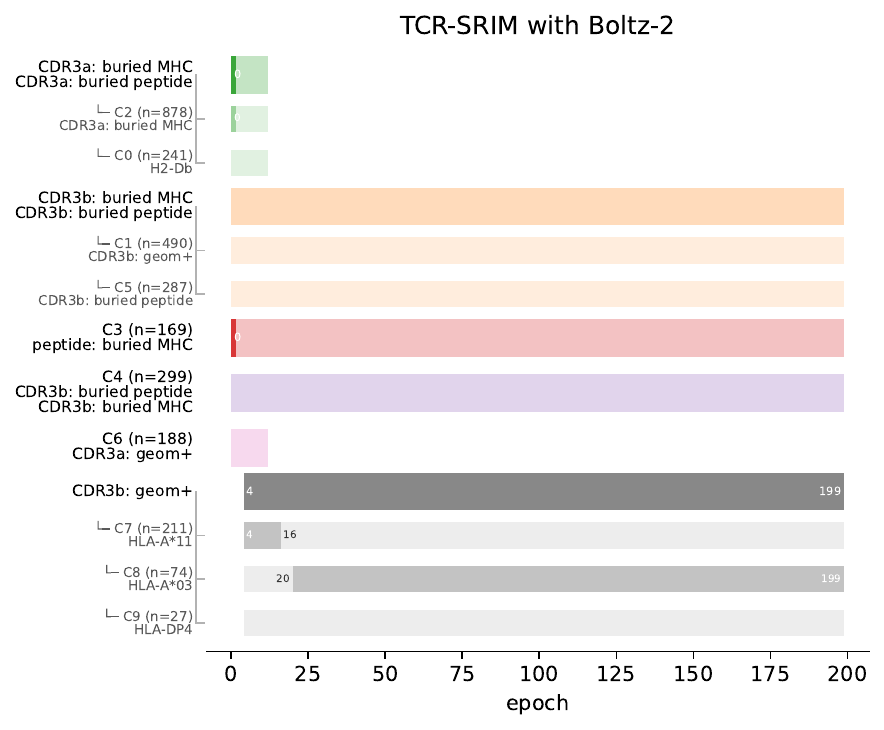}
    \includegraphics[width=0.49\linewidth]{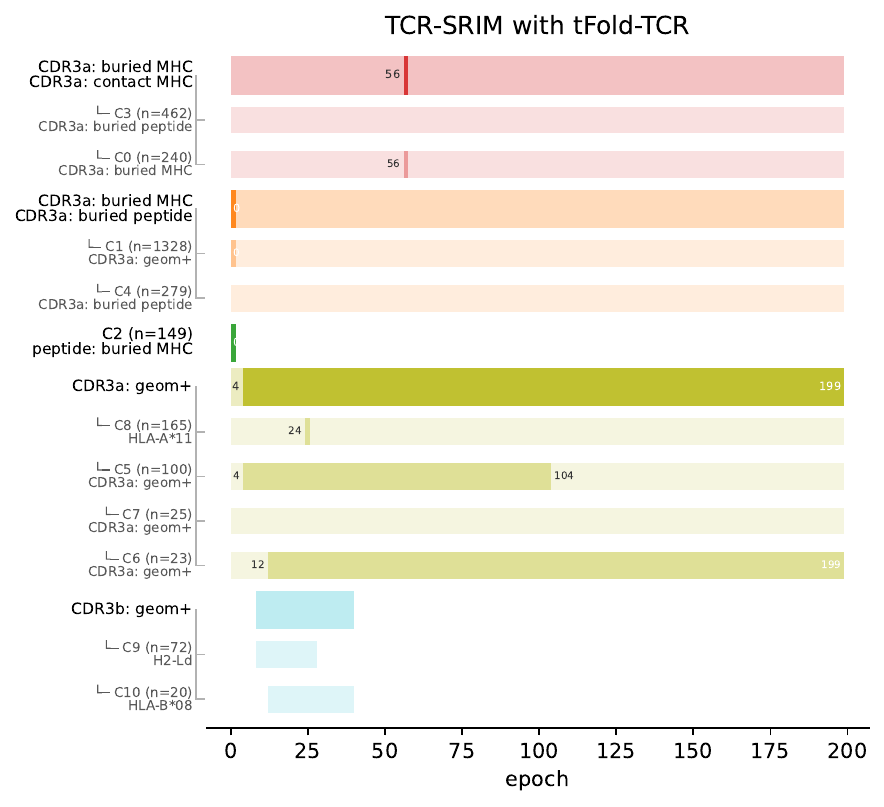}
    \includegraphics[width=0.49\linewidth]{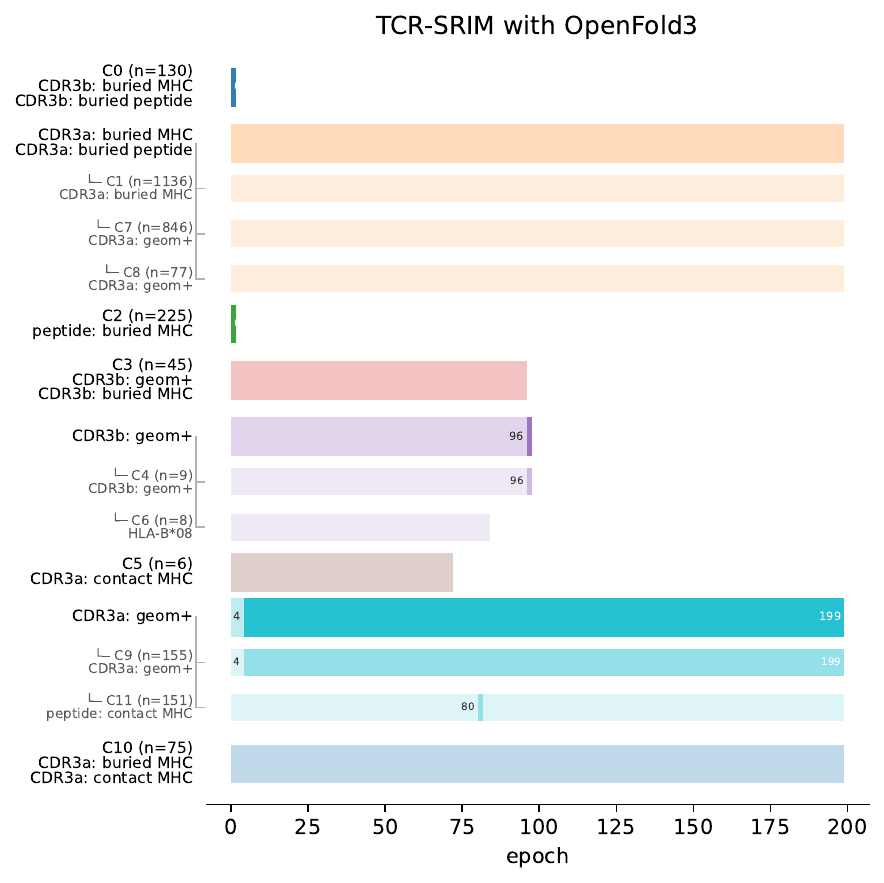}
    \caption{\ModelName[abbr] for TCR-SRIM regularized with experimentally resolved structures and structures predicted by AlphaFold3, Boltz-2, TCRModel2, tFold-TCR, and OpenFold3.}
    \label{fig:srim:eft:full}
\end{figure}

We then regularize TCR-SRIM using either experimentally resolved or predicted structures from the TCR-XAI2 training split and evaluate the evolution of BRHR on the test split using experimentally resolved structures. As shown in Fig.~\ref{fig:srim:brhr:full} and Fig.~\ref{fig:srim:eft:full}, when experimentally resolved structures are used for regularization, the BRHR of all three chains converges to stable values within approximately 70 epochs, with substantially reduced fluctuations. Predicted structures can also stabilize model interpretability, but their effects vary with structural quality.
TCRModel2, which is adapted from AlphaFold2, exhibits a pattern similar to AlphaFold3. However, it enables TCR-SRIM to achieve higher CDR3a-peptide interpretability, reaching a BRHR of approximately 0.7, whereas AlphaFold3 reaches approximately 0.5, similar to the model regularized with experimentally resolved structures. Both AlphaFold-based structures nevertheless stabilize the model's interpretation of the three chains relatively quickly, consistent with their relatively high structural prediction quality.
Although Boltz-2 also achieves high structural prediction quality, it produces a different training trajectory. Compared with the AlphaFold-based models, Boltz-2 leads to higher BRHR for CDR3b-peptide interactions than for CDR3a-peptide interactions, while the latter remains less stable throughout training. For tFold-TCR, which is designed for high-throughput structure prediction, the lower structural quality compared with the AlphaFold-based models and Boltz-2 is accompanied by less stable BRHR trajectories. OpenFold3 produces substantially lower-quality structures than tFold-TCR, and the corresponding model interpretation is also less stable.
Overall, these results suggest that structural information can stabilize the model's understanding of TCR-epitope interactions during training, while the quality of the structural guidance influences both the stability and the interaction-specific patterns learned by the model.

\subsection{Insights for TCR-Epitope Model Design}
According to our analysis of \ModelName[abbr] across different TCR-epitope prediction models, and motivated by the finding that interpretability can provide guidance for model design~\citep{li2025rational}, our results offer several insights for future TCR-epitope prediction models. For models without explicit MHC information, architectural mechanisms that facilitate coordination between the TCR alpha and beta chains, such as cross-attention between the two chains, may help reduce the conflicts observed during training. Another possible strategy is to first learn the pairing between the alpha and beta chains independently before incorporating peptide information, which may encourage the model to establish coordinated TCR representations and reduce subsequent conflicts.

For structure-based models, our results suggest two important considerations. First, incorporating structural information can stabilize the model's interpretation during training. Second, the quality of predicted structures matters for how effectively they guide model learning. One possible direction is therefore to combine structures from different prediction models to provide complementary structural information, for example, using Boltz-2 and AlphaFold-based structures to better capture CDR3b-peptide and CDR3a-peptide interactions, respectively.

\end{document}